\pdfoutput=1
\documentclass[10pt,logo,copyright]{nvidiatechreport}
\usepackage[numbers]{natbib}

\usepackage[utf8]{inputenc} 
\usepackage[T1]{fontenc}    

\usepackage{parskip}        
\usepackage{url}            
\usepackage{hyperref}       
\usepackage{booktabs}       
\usepackage{amsfonts}       
\usepackage{nicefrac}       
\usepackage{microtype}      
\usepackage{xcolor}         
\usepackage[dvipsnames]{xcolor} 

\definecolor{mygreen}{RGB}{60,179,113}
\definecolor{mygray}{RGB}{130 130 130}
\usepackage{graphicx}
\usepackage{subcaption}
\usepackage{pifont}
\usepackage{placeins}
\usepackage{animate}        
\usepackage{subcaption}
\usepackage{tabularx}
\usepackage{makecell}
\usepackage{adjustbox}
\usepackage{setspace}
\newcolumntype{M}[1]{>{\centering\arraybackslash}m{#1}}
\usepackage{float}
\usepackage{tikz}
\usetikzlibrary{positioning,shapes,arrows}
\usepackage{amsmath,amsfonts,bm, bbm,leftindex}
\usepackage{multirow}
\usepackage{comment}
\usepackage{gensymb}
\usepackage{lipsum}
\usetikzlibrary{arrows.meta, positioning, fit}
\usepackage[para]{threeparttable}
\usepackage{tikz}
\usetikzlibrary{tikzmark}

\def\eqref#1{equation~\ref{#1}}

\def\1{\bm{1}}

\DeclareMathAlphabet{\mathsfit}{\encodingdefault}{\sfdefault}{m}{sl}
\SetMathAlphabet{\mathsfit}{bold}{\encodingdefault}{\sfdefault}{bx}{n}

\makeatletter
\let\save@mathaccent\mathaccent
\newcommand*\if@single[3]{%
  \setbox0\hbox{${\mathaccent"0362{#1}}^H$}%
  \setbox2\hbox{${\mathaccent"0362{\kern0pt#1}}^H$}%
  \ifdim\ht0=\ht2 #3\else #2\fi
  }
\newcommand*\rel@kern[1]{\kern#1\dimexpr\macc@kerna}
\newcommand*\widebar[1]{\@ifnextchar^{{\wide@bar{#1}{0}}}{\wide@bar{#1}{1}}}
\newcommand*\wide@bar[2]{\if@single{#1}{\wide@bar@{#1}{#2}{1}}{\wide@bar@{#1}{#2}{2}}}
\newcommand*\wide@bar@[3]{%
  \begingroup
  \def\mathaccent##1##2{%
    \let\mathaccent\save@mathaccent
    \if#32 \let\macc@nucleus\first@char \fi
    \setbox\z@\hbox{$\macc@style{\macc@nucleus}_{}$}%
    \setbox\tw@\hbox{$\macc@style{\macc@nucleus}{}_{}$}%
    \dimen@\wd\tw@
    \advance\dimen@-\wd\z@
    \divide\dimen@ 3
    \@tempdima\wd\tw@
    \advance\@tempdima-\scriptspace
    \divide\@tempdima 10
    \advance\dimen@-\@tempdima
    \ifdim\dimen@>\z@ \dimen@0pt\fi
    \rel@kern{0.6}\kern-\dimen@
    \if#31
      \overline{\rel@kern{-0.6}\kern\dimen@\macc@nucleus\rel@kern{0.4}\kern\dimen@}%
      \advance\dimen@0.4\dimexpr\macc@kerna
      \let\final@kern#2%
      \ifdim\dimen@<\z@ \let\final@kern1\fi
      \if\final@kern1 \kern-\dimen@\fi
    \else
      \overline{\rel@kern{-0.6}\kern\dimen@#1}%
    \fi
  }%
  \macc@depth\@ne
  \let\math@bgroup\@empty \let\math@egroup\macc@set@skewchar
  \mathsurround\z@ \frozen@everymath{\mathgroup\macc@group\relax}%
  \macc@set@skewchar\relax
  \let\mathaccentV\macc@nested@a
  \if#31
    \macc@nested@a\relax111{#1}%
  \else
    \def\gobble@till@marker##1\endmarker{}%
    \futurelet\first@char\gobble@till@marker#1\endmarker
    \ifcat\noexpand\first@char A\else
      \def\first@char{}%
    \fi
    \macc@nested@a\relax111{\first@char}%
  \fi
  \endgroup
}
\makeatother

\definecolor{darkred}{rgb}{0.7, 0.0, 0.0}

\usepackage{pifont}

\usepackage[nameinlink]{cleveref}
\crefname{equation}{Eq.}{Eqs.}
\crefname{figure}{Fig.}{Figs.}
\crefname{section}{Sec.}{Secs.}
\crefname{appendix}{App.}{Apps.}
\crefname{table}{Tab.}{Tabs.}
\crefname{algorithm}{Algo.}{Algos.}
\usepackage{url}
\usepackage{hyperref}

\newcommand{\ours}{EgoScale}

\title{\ours: Scaling Dexterous Manipulation with Diverse Egocentric Human Data}

\author{
\parbox{\textwidth}{
Ruijie Zheng$^{1*}$,
Dantong Niu$^{1,2*}$,
Yuqi Xie$^{1*}$,
Jing Wang$^{1}$,
Mengda Xu$^{1}$,
Yunfan Jiang$^{1}$,
Fernando Castañeda$^{1}$,
Fengyuan Hu$^{1}$,
You Liang Tan$^{1}$,
Letian Fu$^{1,2}$,
Trevor Darrell$^{2}$,
Furong Huang$^{3}$,
Yuke Zhu$^{1\dagger}$,
Danfei Xu$^{1\dagger}$,
Linxi Fan$^{1\dagger}$\\
$^{1}$NVIDIA \quad
$^{2}$University of California, Berkeley \quad
$^{3}$University of Maryland\\
$^{*}$Equal Contribution \quad
$^{\dagger}$Project Lead \\
}
\url{https://research.nvidia.com/labs/gear/egoscale/}
}

\begin{abstract}

Human behavior is among the most scalable sources of data for learning physical intelligence, yet how to effectively leverage it for dexterous manipulation remains unclear.
While prior work demonstrates human-to-robot transfer in constrained settings, it is unclear whether large-scale human data can support fine-grained, high-degree-of-freedom dexterous manipulation.
We present \ours, a human-to-dexterous-manipulation transfer framework built on large-scale egocentric human data.
We train a Vision–Language–Action (VLA) model on over 20,854 hours of action-labeled egocentric human video—more than 20× larger than prior efforts—and uncover a log-linear scaling law between human data scale and validation loss.
This validation loss strongly correlates with downstream real-robot performance, establishing large-scale human data as a predictable supervision source.
Beyond scale, we introduce a simple two-stage transfer recipe: large-scale human pretraining followed by lightweight aligned human–robot mid-training.
This enables strong long-horizon dexterous manipulation and one-shot task adaptation with minimal robot supervision.
Our final policy improves average success rate by 54\% over a no-pretraining baseline using a 22-DoF dexterous robotic hand, and transfers effectively to robots with lower-DoF hands, indicating that large-scale human motion provides a reusable, embodiment-agnostic motor prior. 

\end{abstract}

\begin{document}

\maketitle

\abscontent

\begin{figure*}[t]
    \centering
    \includegraphics[width=1.0\textwidth]{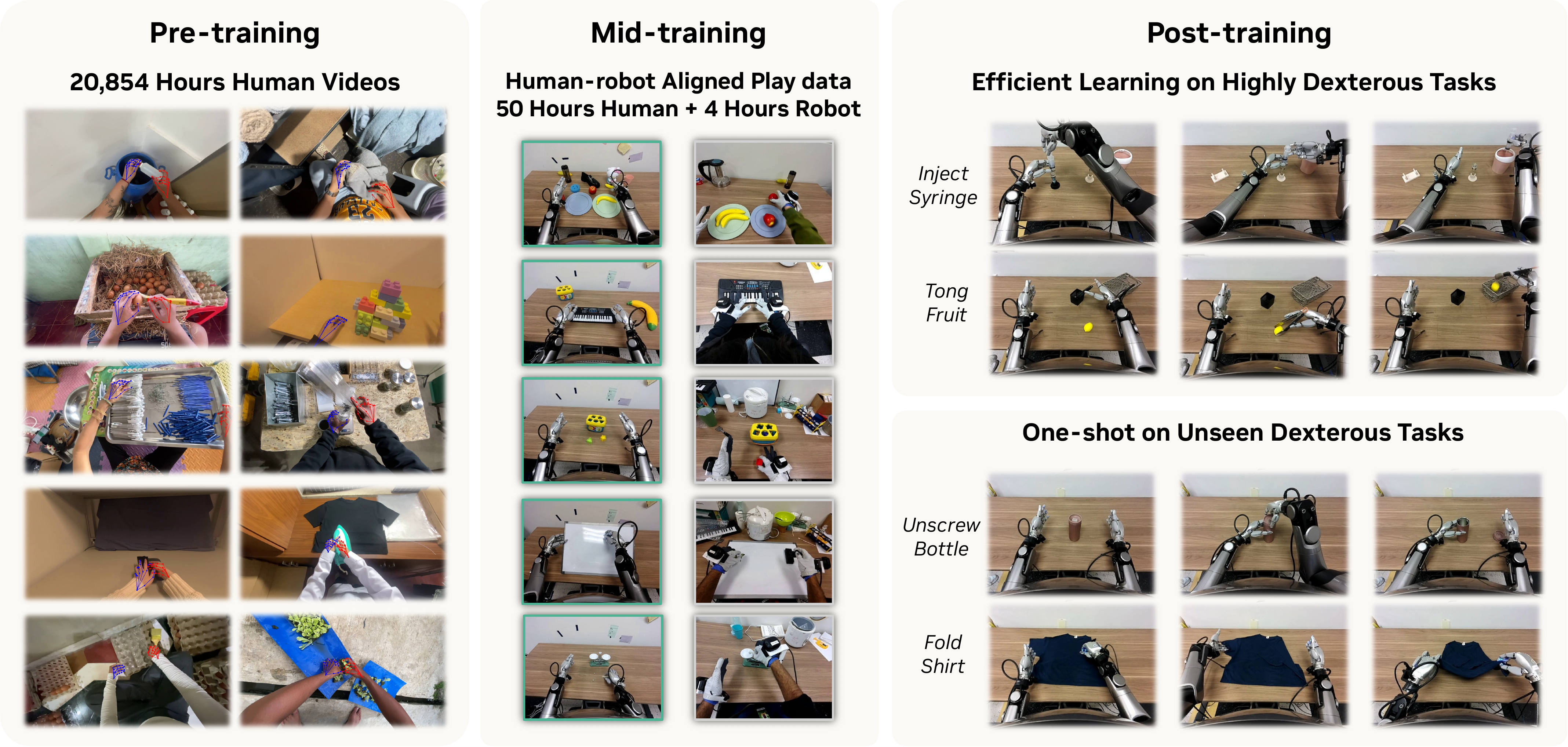}
    \caption{\textbf{\ours: Two-stage human-to-robot learning framework.}
    A flow-based Vision-Language-Action (VLA) policy is first pretrained on 20,854 hours of egocentric human videos using wrist motion and retargeted dexterous hand actions. A lightweight mid-training stage with aligned human robot play data (pairs highlighted with \textcolor{mygreen}{green} and \textcolor{mygray}{gray} boundaries) adapts the representation to robot sensing and control. The resulting policy is post-trained on downstream tasks, enabling efficient learning of dexterous manipulation and one-shot generalization to unseen skills.\looseness=-1}
    \label{fig:method_overview}
\end{figure*}

\section{Introduction}
\label{sec:intro}

Human behavior is one of the most scalable sources of data for learning physical intelligence.
Humans routinely perform dexterous manipulation across diverse objects, environments, and task variations at a scale that far exceeds what can be collected through robot teleoperation.
As robotic hardware continues to improve toward more human-like kinematics and dexterity, a natural question arises: \emph{can human data serve as a primary training signal for dexterous robot manipulation?}

Recent work shows that transfer from human data to robots is possible by aligning observations or actions across embodiments~\cite{kareer2024egomimicscalingimitationlearning, yang2025egovla, qiu2025humanoid, punamiyaegobridge, tao2025dexwild}.
However, existing results remain limited in two respects. First, most approaches rely on relatively small human datasets, typically on the order of tens to hundreds of hours. Second, many focus on grippers or low DoF hands, where fine-grained finger articulation is absent. It therefore remains unclear whether human data can meaningfully support complex, dexterous manipulation at scale.

In this work, we show that human-to-robot transfer for dexterous manipulation is fundamentally a scaling phenomenon, and present~\ours, a scalable human-to-dexterous-manipulation transfer framework built on large-scale egocentric human data.
We pretrain on \textbf{20,854 hours of egocentric human manipulation data}, over \textbf{20$\times$ larger} than the dataset sizes used in prior studies of human–robot policy transfer, and uncover a clear \emph{scaling law}: human wrist and hand action prediction validation loss follows a log-linear relationship with data volume. This enables us to extrapolate: as human data scales, validation loss continues to decrease, and the learned representations generalize increasingly well. Crucially, the loss strongly correlates with real robot performance on long-horizon, complex manipulation tasks. Together, these results establish large-scale human data as a scalable and predictable supervision source for learning dexterous manipulation policies. 

Beyond scale, we identify a simple yet effective training recipe that enables new generalization capabilities.
We supervise the model using human manipulation behaviors represented as relative wrist motion and retargeted high-DoF hand joint actions.
This aligned action space encourages the model to extract information that is directly useful for manipulation, rather than learning task-agnostic visual features.
After pretraining, we introduce a small amount of aligned human-robot mid-training data through co-training. 
The mid-training data includes humans and robots performing similar manipulation tasks in matched tabletop scenes with comparable visual viewpoints.
This alignment provides supervisions for grounding the pretrained representations in the robot’s sensing and control space. \looseness=-1

Importantly, this mid-training stage gives rise to \emph{emergent one-shot and few-shot generalization}.
With only one or a few robot demonstrations, the policy can adapt to new dexterous tasks without requiring extensive task-specific data collection.
For instance, using a single robot demonstration, the trained policy achieves up to \textbf{88\% average success} on shirt folding, even though the mid-training data contains only folding behaviors.
Moreover, although human actions are supervised in a high-DoF dexterous hand space, the learned representations generalize to \emph{substantially different robot embodiments}.
On the Unitree G1 robot with a tri-finger hand, human-pretrained policies also achieve over \textbf{30\% absolute improvement} in success rate across both evaluated tasks compared to baseline without human pretraining.


Taken together, our results show that effective dexterous human-to-robot transfer requires scale, explicit motion supervision, and a small amount of precise human–robot alignment.
Rather than replacing robot data, large-scale human demonstrations dramatically amplify its effectiveness, pointing toward a future where humans can be treated as another scalable embodiment in robot learning. We summarize the contributions as follows:

\begin{itemize}
    \item \textbf{Human Data Pretraining Scaling Laws.}
At the scale of over 20k hours of human data, we uncover a clear log-linear relationship between data scale and hand-action prediction loss and show that this loss strongly correlates with real-robot dexterous manipulation performance.

\item \textbf{An Effective Human-to-Robot Transfer Recipe.} We combine high-DoF human hand action supervision with a small amount of aligned human–robot mid-training, enabling strong post-training performance with minimal robot data.

\item \textbf{Emergent One-shot Transfer and Generalization.} Our approach enables one-shot transfer to previously unseen dexterous task with 22-DoF hands. 
The pretraining policy also generalizes effectively to robots with lower-DoF hands, indicating that rich human motion provides a reusable motor prior.
\end{itemize}

\section{Method}
\label{sec:model}

We aim to learn representations from large-scale egocentric human video that are directly useful for dexterous robot control. This setting poses two core challenges. First, human demonstrations are noisy and lack paired robot actions. Second, human and robot embodiments differ substantially in kinematics and control interfaces. Our method (Figure~\ref{fig:method_overview}) addresses these challenges through two design choices. 
We first pretrain on human data using explicit supervision of wrist motion and hand articulation extracted from egocentric videos, forcing the model to learn physically grounded action representations.
We then introduce a small amount of aligned human-robot data for mid-training, which grounds these representations in executable robot control without requiring large-scale paired demonstrations. Together, this two-stage design decouples data scale from embodiment alignment, enabling effective transfer from large human datasets to dexterous robot manipulation. 



\subsection{Human Action Representation}
\label{sec:action_representation}

\textbf{Raw Sensor Streams.}
Each human demonstration consists of egocentric RGB observations captured from a head-mounted camera, together with estimated camera motion and human hand pose obtained from off-the-shelf perception pipelines. We convert these raw sensory signals into a unified action representation suitable for large-scale pretraining and downstream robot execution. Let $\mathcal{F}_w$ denote the world frame and $\mathcal{F}_c^t$ the camera frame at time $t$. The estimated camera pose is represented as $\mathbf{T}_{w \leftarrow c}^t \in \mathbb{SE}(3)$. Human hand pose is modeled by 21 keypoints, each represented as a rigid transform $\mathbf{H}_{c,i}^t \in \mathbb{SE}(3)$ in the camera frame, where $i=1$ corresponds to the wrist. The wrist pose in the world frame is given by $\mathbf{W}_w^t = \mathbf{T}_{w \leftarrow c}^t \mathbf{H}_{c,1}^t$.

\textbf{Wrist-level Arm Motion.}
To obtain motion commands that are invariant to global camera movement, we represent arm motion using relative wrist motion between consecutive timesteps. 
Given timestep $t$ in an action chunk, $\Delta \mathbf{W}^t = (\mathbf{W}_w^{0})^{-1} \mathbf{W}_w^t$. This relative end-effector formulation removes dependence on absolute camera pose and captures local arm motion in a physically meaningful manner. The same representation is shared across human demonstrations and robot executions, serving as the primary arm-level action abstraction for cross-embodiment learning.

\textbf{Hand Articulation.}
For finger-level control, we retarget the 21 human hand keypoints into a dexterous robot hand joint space using an optimization-based procedure that enforces joint limits and kinematic constraints.
Our default choice is the 22-DoF hand action space of the Sharpa hand~\cite{sharpa_wave}, which preserves human finger articulation during pretraining while aligning with the control interface of our target robot.
Although this representation is defined using a high-DoF hand, we later show that the learned models transfer effectively to robots with lower-DoF hands.


\subsection{Human Data Sources and Processing}
\label{sec:human_data}

\noindent\textbf{Stage I: Large-Scale Egocentric Human Pretraining Data.}
We pretrain our model on a large-scale mixture of egocentric human activity datasets totaling \textbf{20,854} hours of video.
The majority consists of in-the-wild egocentric recordings spanning diverse real-world environments (e.g., household, industrial, retail, and educational settings), covering 9,869 scenes, 6,015 tasks, and 43,237 objects, and providing broad coverage of long-tailed manipulation behaviors.

\begin{figure}[!htbp]
    \centering
    \begin{subfigure}[t]{0.49\linewidth}
        \centering
        \includegraphics[width=\linewidth]
        {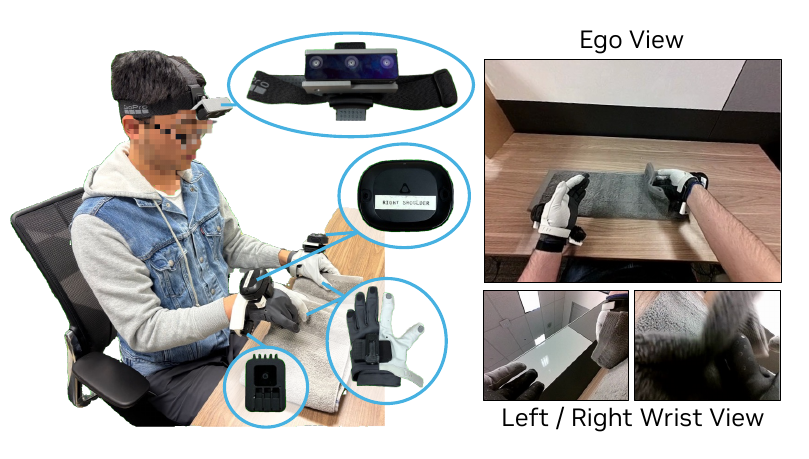}
        \caption{Egocentric Human Data Collection.}
        \label{fig:ego_collection}
    \end{subfigure}
    \hfill
    \begin{subfigure}[t]{0.49\linewidth}
        \centering
        \includegraphics[width=\linewidth]
        {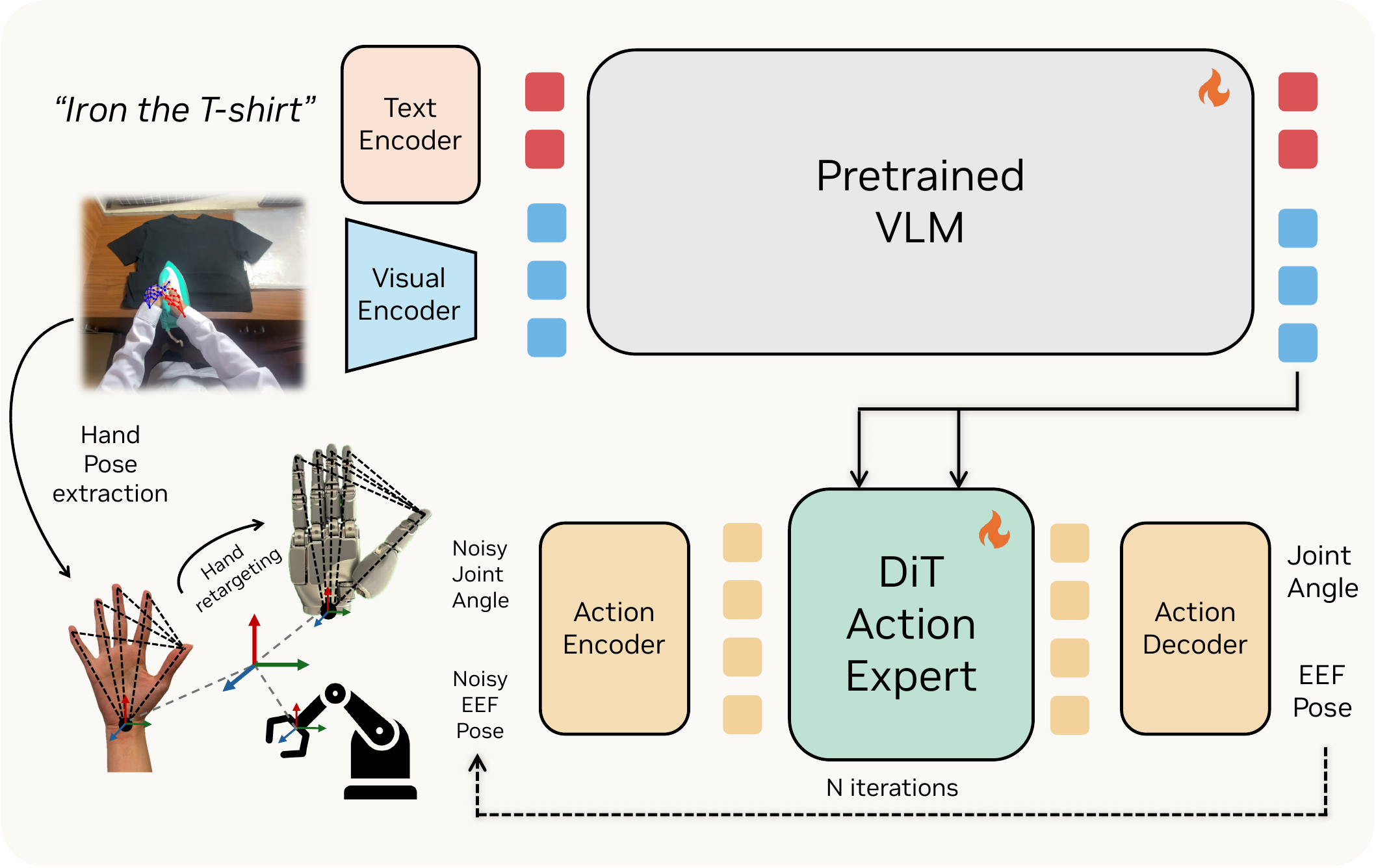}
        \caption{\ours~Model Architecture.}
        \label{fig:arch}
    \end{subfigure}
    
    \caption{
    \textbf{Human Data Collection and Model Architecture.}
    (\textbf{Left}) Aligned human-robot mid-training data are collected using the same sensing setup as the robot. 
    Vive trackers and Manus gloves capture arm and hand motion, while one head-mounted camera and two wrist-mounted cameras record egocentric and wrist views, enabling consistent perception–action alignment. 
    (\textbf{Right}) A flow-based VLA policy with a VLM backbone and DiT action expert. 
    Human and robot data are unified through a wrist-level action representation, with lightweight embodiment-specific adapters for proprioception and hand actions.
    }
    \label{fig:overview}
    \vspace{-5pt}
\end{figure}


All recordings are captured using egocentric RGB cameras at 30 FPS.
We apply off-the-shelf SLAM and hand-pose estimation pipelines to recover camera motion and human hand trajectories.
Although these estimates are noisy due to unconstrained data collection, the scale and diversity of the data provide effective supervision for learning transferable action representations, which continue to improve downstream performance as data volume increases.

To complement this large-scale but noisy supervision, we additionally incorporate 829 hours of EgoDex dataset~\cite{hoque2025egodex}, collected using Apple Vision Pro with accurate wrist and hand tracking.
EgoDex covers 194 tabletop manipulation tasks involving everyday objects and provides higher-precision kinematic signals that help anchor pretraining while preserving scalability. \looseness=-1




\noindent\textbf{Stage II: Aligned Human-Robot Mid-Training Data.}
To further bridge the embodiment gap between human demonstrations and robot execution, we introduce a smaller dataset with both human and teleoperated robot data. We later show that this dataset is critical to anchor the pretrained representations to the robot’s sensing and action spaces.

This dataset comprises 344 tabletop manipulation tasks, with each task captured in approximately 30 human trajectories and 5 robot trajectories, totaling about 50 hours of human data and only 4 hours of robot data. As shown in Figure~\ref{fig:ego_collection}, human demonstrations are collected using the same camera configuration as the robot, with matched viewpoints and calibrated intrinsics, ensuring that visual observations are directly comparable across domains.
Human hand motion is captured using the same motion-capture stack as in robot teleoperation: Vive trackers provide wrist pose (3D position and orientation), while Manus gloves record full in-hand pose as 25 joint transforms.
All motion signals are synchronized with the video stream.

Compared to the large-scale but unconstrained data used in Stage~I, this dataset is significantly smaller but explicitly embodiment-aligned.
It focuses on tabletop tasks that match the robot’s workspace and kinematics, enabling abstract human actions learned during pretraining to be grounded in executable robot control. Together, Stage~I and Stage~II decouple scale and alignment: Stage~I provides diversity and semantic grounding, while Stage~II supplies precise human-robot correspondence for downstream deployment.


\subsection{Model Architecture}
\label{sec:models}
As shown in Figure~\ref{fig:arch}, our model follows a flow-based VLA architecture similar to GR00T N1~\cite{nvidia2025gr00tn1openfoundation}.
At each timestep $t$, the model conditions on an observation $o_t = (I_t, l_t)$ consisting of an image and a language instruction, which is encoded into a vision-language embedding $\phi_t$.
The model then predicts a chunk of future actions using a flow-matching objective.

For robot data, the model conditions on the robot proprioceptive state $q_t$, while human demonstrations do not provide such signals.
In the absence of proprioception, we replace $q_t$ with a learnable placeholder token, enabling a unified model formulation without architectural changes.
To accommodate multiple robot embodiments with different state and hand action spaces, following GR00T N1~\cite{nvidia2025gr00tn1openfoundation}, we use lightweight embodiment-conditioned MLP adapters at the input and output interfaces.
Specifically, these adapters encode embodiment-specific proprioceptive state and decode hand actions, while relative wrist motion prediction, the vision-language backbone, and the DiT action expert are fully shared.
In practice, this mechanism is used only for a small number of additional embodiments (e.g., G1 with a tri-finger hand).

\subsection{Training Recipe}
\label{sec:training}

We use a three-stage training pipeline.
In \textbf{Stage I} (human pretraining), we train on 20K hours of egocentric human data for 100K steps with 256 GB200 GPUs using a global batch size of 8{,}192 and learning rate $5\times10^{-5}$, fully unfreezing every parameter of the VLA model to absorb large-scale data.
Then in \textbf{Stage II} (aligned mid-training), we train on the aligned human-robot play dataset for 50K steps with batch size 2{,}048 and learning rate $3\times10^{-5}$, freezing the vision-language backbone while only updating the vision encoder and DiT action expert to anchor representations to robot sensing and control.
In \textbf{Stage III} (post-training), we fine-tune on task-specific robot demonstrations for 10K steps with batch size 512 and learning rate $3\times10^{-5}$. 
During post-training, the vision encoder is frozen if mid-training is used and unfrozen otherwise, to accommodate new embodiments when needed.

\subsection{Robot Systems and Control}
\label{sec:systems}

The real world experiments are conducted on the Galaxea R1Pro humanoid robot with 22-DoF Sharpa dexterous robot hands. 
We refer the readers to Appendix B for the system figure. \looseness=-1

\noindent\textbf{Dual-arm Wheeled Humanoid System Galaxea R1Pro.} 
We fix the base and torso and focus on bimanual manipulation, controlling both 7-DoF arms in relative end-effector space where actions specify incremental position and orientation changes, matching the wrist-pose representation used in human demonstrations for direct human-robot alignment.

\noindent\textbf{22-DoF Dexterous Hands.} 
We equip the robot with Sharpa Wave hands with 22 degrees of freedom and joint-space control, where actions directly specify target joint angles, enabling precise articulation and preserving the fine-grained structure of retargeted human hand motion.

\noindent\textbf{Perception System.}
We use three RGB cameras: a head-mounted camera that provides an egocentric first-person view consistent with human videos, and two wrist ones mounted on the inner side of each wrist facing the palm, capturing close-range hand-object interactions and provide detailed visual feedback essential for fine-grained dexterous manipulation.


\section{Experiment}
\label{sec:exp}
In this section, we aim to answer the following research questions through our experiments:\\
\noindent\textbf{RQ1:} \emph{Does large-scale egocentric human pretraining improve downstream dexterous manipulation performance compared to training from scratch or embodiment-aligned data alone?}\newline
\noindent\textbf{RQ2:} \emph{How does the scale of human pretraining data affect representation quality and real-robot performance?}\newline
\noindent\textbf{RQ3:} \emph{What role does mid-training play in enabling few-shot adaptation and generalization to novel tasks\looseness=-1?}\newline
\noindent\textbf{RQ4:} \emph{Do human-pretrained representations transfer across robot embodiments with substantially different kinematics and control interfaces?}\newline
\noindent\textbf{RQ5:} \emph{How does the choice of human action representation during pretraining affect downstream dexterous manipulation?}

\subsection{Experiment Setup}

\begin{figure}[!htbp]
    \centering
    \includegraphics[width=\linewidth]{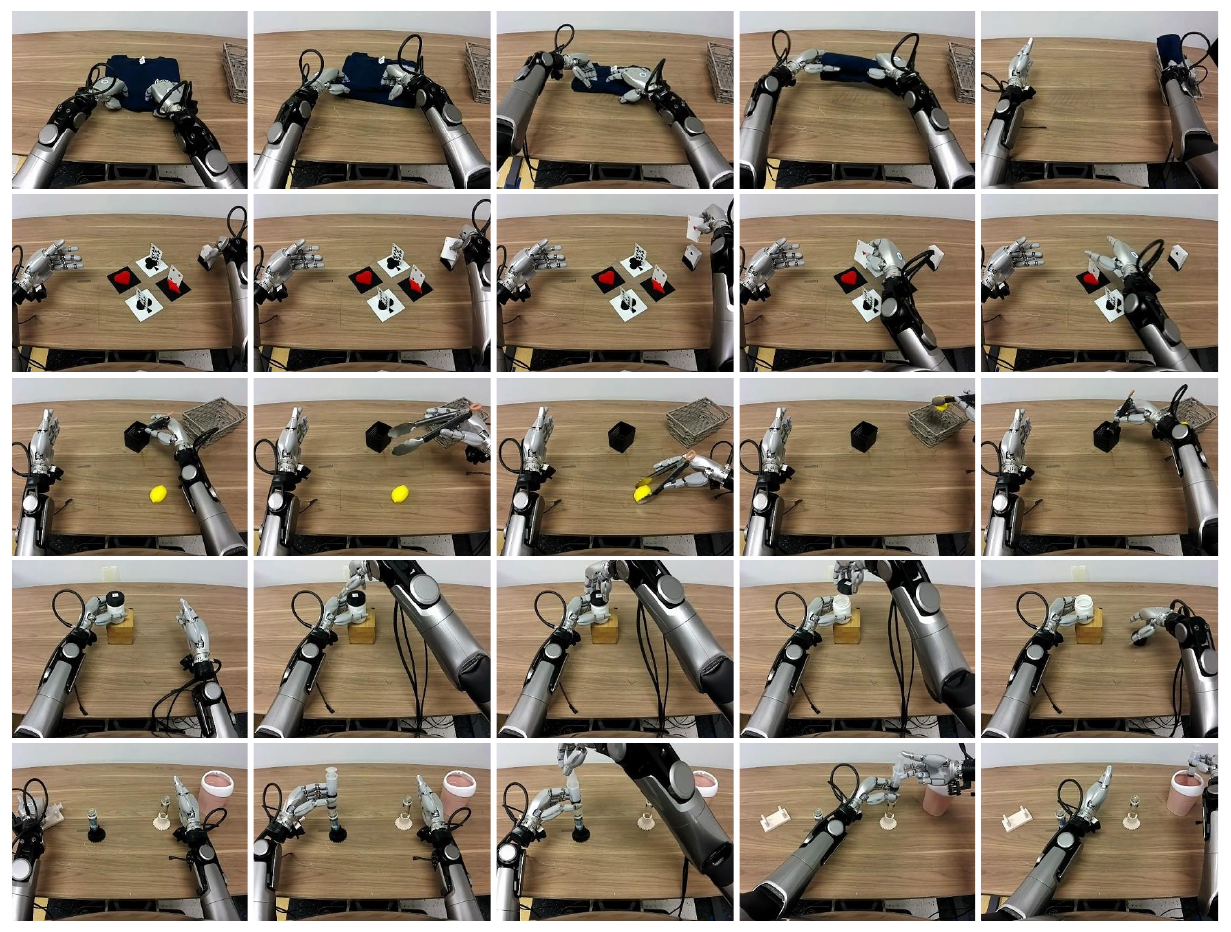}
    \caption{\textbf{Post-Training Evaluation Tasks.} 
Five dexterous manipulation tasks used to evaluate post-training performance
}
    \label{fig:task_fig}
\end{figure}

\noindent\textbf{Tasks.}
To evaluate policy performance, we design five highly dexterous manipulation tasks shown in Figure~\ref{fig:task_fig}. 

Each task is provided with 100 teleoperated robot demonstrations, except for \emph{Shirt Rolling}, a deformable manipulation task that requires less precise control, for which we use only 20 demonstrations.

\textbf{(Task I)} \emph{(Shirt) Shirt Rolling.}
The robot coordinates both hands to alternately fold and roll a T-shirt into a cylindrical shape before placing it into a basket.

\textbf{(Task II)} \emph{(Card) Card Sorting.}
The robot uses its fingers to rub and separate a single card from a tightly stacked deck, and then precisely inserts it into the correct holder based on color.

\textbf{(Task III)} \emph{(Tong) Dexterous Tool Use: Tongs for Fruit Transfer.}
The robot first grasps a pair of tongs from a toolbox and then uses them to pick up a fruit and place it at a target location.

\textbf{(Task IV)} \emph{(Bottle) Unscrewing a Bottle Cap.}
The robot grasps and continuously rotates a small cap to remove it from a bottle.
We collect demonstrations on four bottles of different sizes, with 25 trajectories per bottle.

\textbf{(Task V)} \emph{(Syringe) Syringe Liquid Transfer.}
This is the most challenging task, requiring the robot to pick up a syringe, draw liquid from tube A, inject it into tube B, and discard the syringe into a trash can.
The task involves long-horizon, multi-step reasoning, precise spatial alignment for fluid extraction and injection, and dexterous manipulation of the syringe plunger.

\noindent\textbf{Evaluation Metric.}
To evaluate policy performance, we train each method using two random training seeds. 
Then, for each trained policy checkpoint, we evaluate performance over 10 trials, except for Task III, where we conduct 4 trials per bottle across four bottle instances, resulting in 16 evaluation trials.
To ensure consistency across evaluation runs, we employ an image-overlay–based initialization procedure, in which the robot evaluator is provided with a visual overlay of the target initial scene configuration to reduce variability in initial conditions.
For each task, we record both the absolute task success rate and fine-grained task completion score.

\subsection{Large-Scale Human Pretraining Is Key to Strong Dexterous Manipulation Policy Performance}
To evaluate the impact of large-scale human pretraining and aligned mid-training on policy learning efficiency, we compare four checkpoints:
(1) a model trained from scratch,
(2) a model pre-trained only on the midtrained aligned human–robot play dataset,
(3) a model pretrained on large-scale human data, and
(4) a human-pretrained model further mid-trained on aligned human–robot data.
For each checkpoint, we report both the task completion score and the absolute success rate.

\begin{figure*}[!t]
    \centering
    \vspace{-0.5em}
    \includegraphics[width=\textwidth]{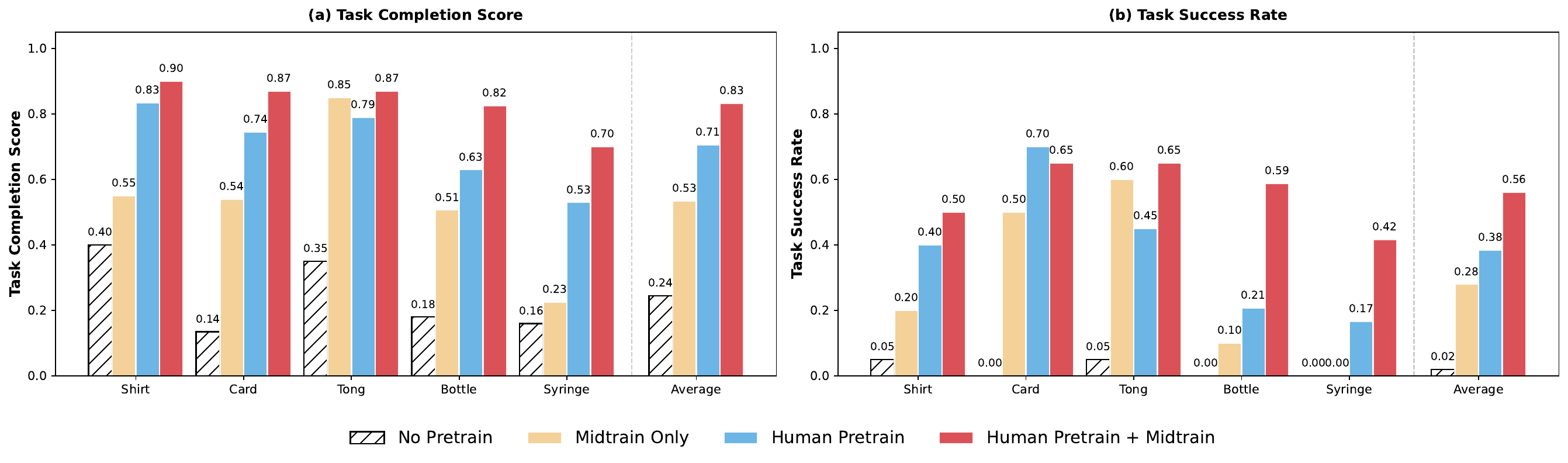}
    \caption{\textbf{Main Experimental Results.} 
Comparison of Human Pre-train + Mid-Training, Human Pretraining, and No Pretraining across five dexterous manipulation tasks under two evaluation metrics.}
\label{fig:main_experimental_results}
\end{figure*}

\begin{figure*}[htbp!]
    \centering
    \includegraphics[width=1\textwidth]{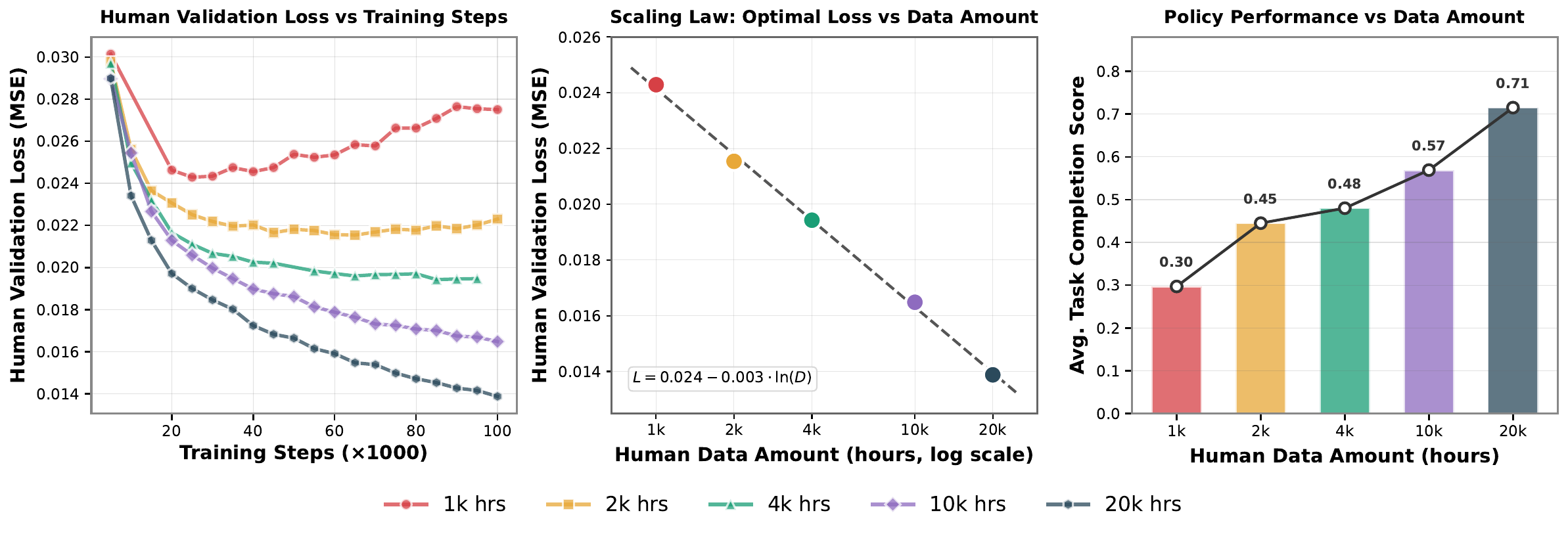}
    \vspace{-0.5em}
    \caption{\textbf{Scaling behavior of human pretraining.}
    \emph{Left:} Human validation loss versus training steps for models pretrained with increasing amounts of egocentric human data (1k–20k hours).
    Larger datasets yield stable, monotonic improvements, while smaller datasets exhibit early overfitting.
    \emph{Center:} Optimal validation loss at convergence as a function of human data scale, revealing a near-perfect log-linear scaling law ($R^2=0.9983$).
    \emph{Right:} Downstream robot performance after post-training, measured by average task completion score, improves consistently with increased human data scale.
    Together, these results demonstrate predictable scaling of learned action representations and their direct translation to improved dexterous manipulation performance.\looseness=-1}
    \vspace{-0.5em}
    \label{fig:human_scaling}
\end{figure*}

Results are summarized in Figure~\ref{fig:main_experimental_results}.
Across all tasks, human pretraining consistently yields substantial performance gains over training from scratch, improving average task completion by over 55\%.
Notably, large-scale human pretraining, despite being noisy, unconstrained, and not task- or sensor-aligned, already outperforms the mid-training-only baseline across most tasks.
This provides the evidence that scale and diversity of human demonstrations provide strong inductive biases for dexterous manipulation, even in the absence of precise embodiment alignment.

Finally, combining human pretraining with a small amount of aligned mid-training yields the best overall performance, suggesting a complementary effect: large-scale human data supplies general manipulation structure, while mid-training anchors these representations to executable robot control.

\subsection{Policy Performance Scales with Pretraining Data Size}
We study how the scale of egocentric human pretraining data affects downstream \emph{real-robot} manipulation performance, and analyze how this behavior is reflected in offline human-action prediction metrics.
We pretrain models using \textbf{1k, 2k, 4k, 10k, and 20k hours} of human data.
To isolate the effect of stage II midtraining, we directly post-train each checkpoint on the downstream task and evaluate their performance.

As shown in Figure~\ref{fig:human_scaling} (right), increasing the amount of human pretraining data leads to consistent and substantial gains in downstream robot performance.
Average task completion rises monotonically from 0.30 at 1k hours to 0.71 at 20k hours, with no signs of saturation in the explored regime.
These results indicate that large-scale human data provides increasingly strong priors for dexterous manipulation, even though the human demonstrations are noisy, unconstrained, and not task-aligned.\looseness=-1

To better understand this trend, we examine how pretraining data scale affects the quality of learned action representations.
We evaluate each pretrained model on a held-out human-video validation set consisting of 2,000 egocentric episodes.
For evaluation, we randomly sample 20 timesteps per trajectory; at each timestep, we draw 16 samples from the flow-matching policy, average the predicted action chunks, and compute the mean squared error against ground-truth wrist and hand actions. \looseness=-1

Figure~\ref{fig:human_scaling} (left) shows human validation loss as a function of training steps for different data scales.
Models trained with smaller datasets (1k--2k hours) initially reduce validation loss but later plateau or degrade, indicating overfitting to limited behavioral diversity.
In contrast, models trained with larger datasets (10k--20k hours) exhibit stable, monotonic improvement throughout training without signs of overfitting.

Strikingly, when we plot the optimal validation loss achieved at convergence against data scale (Figure~\ref{fig:human_scaling}, center), we observe a remarkably clean log-linear scaling law:
\begin{equation}
\mathbf{
L = 0.024 - 0.003 \cdot \ln(D)},
\end{equation}
where $D$ denotes the number of hours of human pretraining data.
The fitted curve achieves an $R^2$ of \textbf{0.9983}, indicating an almost perfect linear relationship in log space.
Crucially, this offline scaling behavior is strongly predictive of real-robot performance.
Human validation loss closely tracks downstream task completion across data scales, establishing it as a meaningful indicator of embodied control capability rather than a purely offline metric.

Together, these results show that effective human-to-robot transfer for dexterous manipulation is fundamentally a scaling phenomenon.
Within the explored regime, increasing human data yields predictable reductions in validation loss and corresponding improvements in robot performance, with no evidence of diminishing returns.
While we do not extrapolate beyond the measured range, this trend suggests substantial headroom for further gains as both human data scale and model capacity continue to increase.

\subsection{Aligned Human-Robot Mid-Training Enables One-Shot Transfer}
\begin{figure*}[!htbp]
    \vspace{-0.5em}
    \centering
    \includegraphics[width=\linewidth]{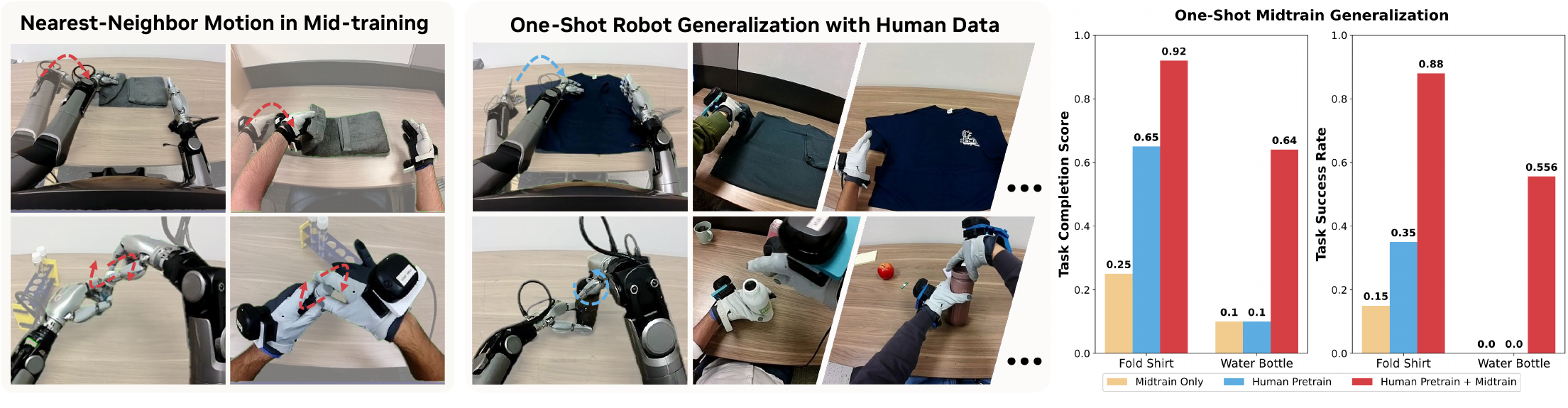}
    \caption{\textbf{Aligned mid-training enables emergent one-shot transfer.}
During post-training, the policy is trained on only a single robot demonstration per task, together with aligned human demonstrations (100 trajectories per object).}
    \label{fig:midtrain_exp}
    \vspace{-0.5em}
\end{figure*}




We evaluate whether aligned human–robot mid-training enables learning of previously unseen skills under extremely limited robot supervision.
Starting from a human-pretrained model, we apply mid-training on the aligned play dataset and post-train on new tasks using only a single robot demonstration, supplemented by aligned human demonstrations.
We consider two tasks, \emph{Fold Shirt} and \emph{Unscrewing Water Bottles}, neither of which appears in the mid-training data.

For \emph{Fold Shirt}, we provide one robot demonstration and 100 aligned human demonstrations.
For \emph{Unscrewing Water Bottles}, we evaluate generalization across object variation using three bottles of different geometries, with one robot demonstration and 100 aligned human demonstrations per bottle.

As shown in Figure~\ref{fig:midtrain_exp}, models that omit either large-scale human pretraining or aligned mid-training fail in this one-shot setting.
In contrast, the \textbf{Pretrain + Midtrain} model achieves success rates of 0.88 on \emph{Fold Shirt} and 0.55 on \emph{Unscrewing Water Bottles}, demonstrating strong few-shot generalization.
Failures are typically partial: incomplete folds for shirts, or unstable grasp maintenance after cap removal for bottles.
These results indicate that aligned mid-training enables a form of transfer that does not emerge from human pretraining or embodiment-specific data alone.

This generalization is enabled by shared motion structure between mid-training and evaluation tasks.
Although the objects and task instances differ substantially, the mid-training data exposes the model to common motion primitives, allowing these behaviors to transfer with as little as a single target-robot demonstration. \looseness=-1

\subsection{Human Pretraining Enables Cross-embodiment Transfer}

\begin{figure}[!htbp]
    \centering
    \includegraphics[width=0.7\linewidth]{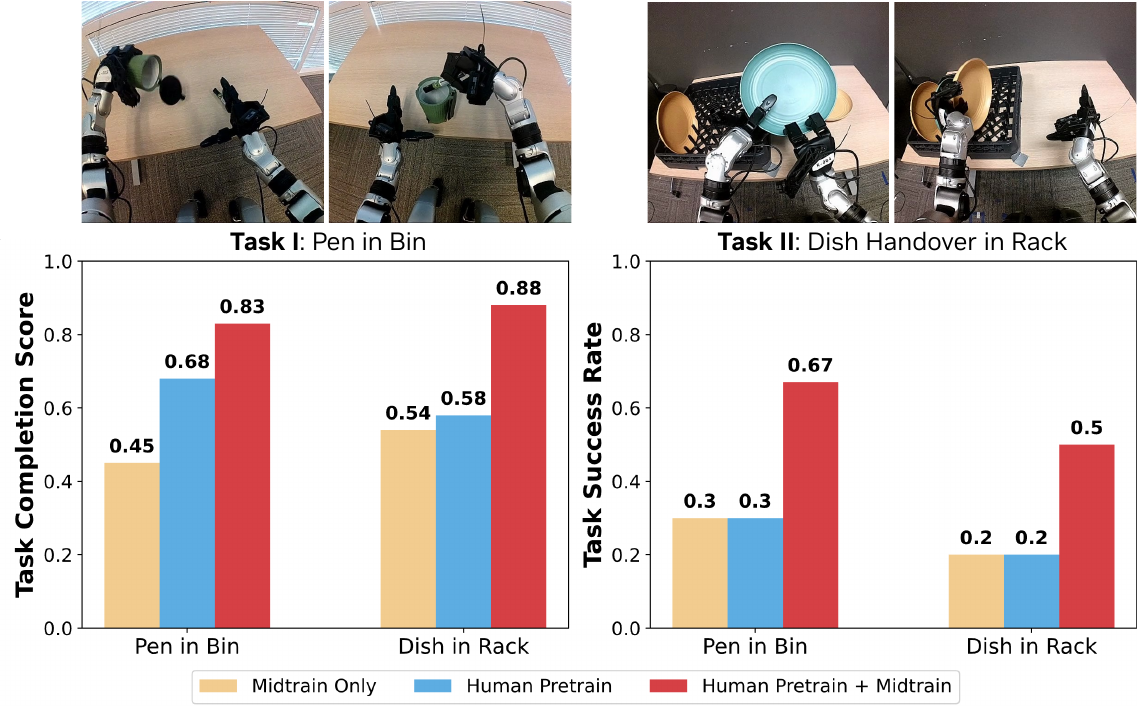}
    \caption{
    \textbf{Mid-training enables cross-embodiment generalization to G1 Tasks.} 
    Integrating a small amount of G1 play data during mid-training enables strong transfer to a low-DoF tri-finger hand, outperforming mid-training or pre-training only.
    }
    \vspace{-0.5em}
    \label{fig:g1_exp}
\end{figure}
Beyond bimanual tabletop manipulation with high-DoF dexterous hands, we also demonstrate that human pretraining learns transferable action representations that generalize to new robot embodiments.
Our human pretraining represents actions as relative $\mathbb{SE}(3)$ end-effector motions coupled with a 22-DoF dexterous hand joint space defined by the Sharpa hand. 
Although this action space is instantiated using a specific robotic hand, it could also function as a motor prior that encodes reusable grasping and manipulation primitives, such as hand opening, closure, and coordinated finger articulation, that is transferable to other robotic hands.
To evaluate cross-embodiment generalization, we introduce a substantially different robot platform: the Unitree G1, which features a shorter arm with a reduced reachable workspace and a 7-DoF tri-finger hand, resulting in markedly different kinematics.

We evaluate posttraining policy performance on two tasks.
In the first task, \emph{Pen in Bin}, the G1 robot uses its left arm to open a trash bin and then uses its right hand to pick up a pen and place it into the bin.
In the second task, \emph{Dish in Rack}, the robot is presented with three plates on a table and must perform a right-to-left handover to place the dishes onto a rack.
During execution, the policy predicts upper-body target commands, while lower-body balance and locomotion are handled by a separately trained Homie~\cite{ben2025homiehumanoidlocomanipulationisomorphic} policy that outputs lower-body joint commands.
Same as in previous experiments, we train each method using two random seeds. For each seed, we evaluate the resulting policy over 10 trials and report the average policy score and success rate across the two seeds.

As shown in Figure~\ref{fig:g1_exp}, incorporating mid-training on a data mixture that includes G1 embodiment play data also leads to a substantial performance improvement on both of the G1 manipulation tasks, compared to a checkpoint trained on the same G1 embodiment–specific data alone. 
Crucially, this improvement cannot be attributed to increased exposure to G1 trajectories alone: policies pretrained and fine-tuned directly on the same G1 dataset, without prior human pretraining, fail to achieve comparable success rates. 
Empirically, we also observe that policies pretrained on human data exhibit qualitatively smoother behaviors.

These results suggest that ~\ours pre-training learns a reusable manipulation structure that transfers across embodiments, while mid-training adapts this structure to the G1’s sensing and control interfaces.

\subsection{Hand Action Space Design for Human Pretraining}
\begin{figure}[!htbp]
    \centering
    \vspace{-0.5em}
    \includegraphics[width=0.7\columnwidth]
    {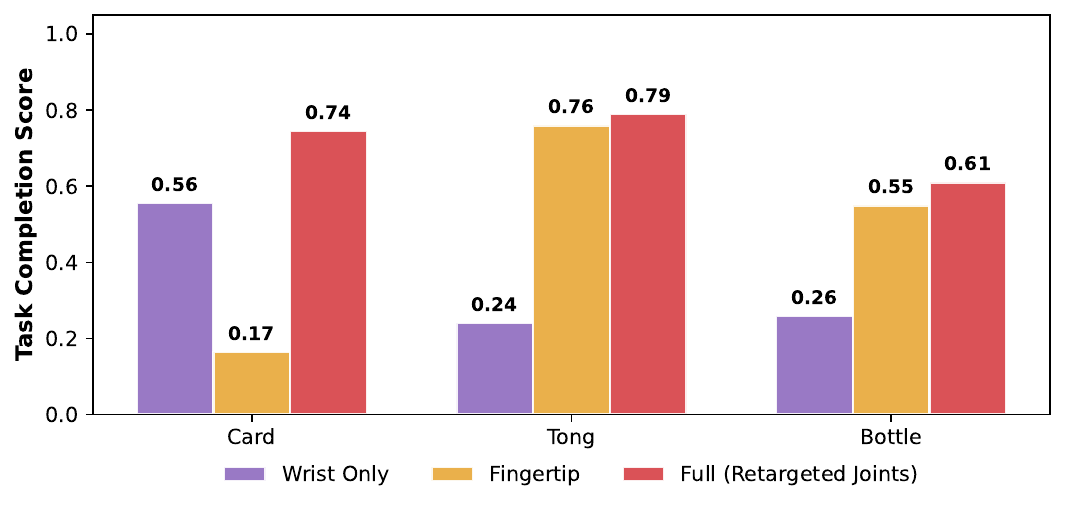}
    \vspace{-0.8em}
    \caption{Task completion score across tasks for different human pretraining action representations.}
    \label{fig:action_space_ablation}
    \vspace{-0.5em}
\end{figure}

We study how the choice of human hand action representation during pretraining affects downstream dexterous manipulation.
Our default setup represents human hand motion using a 22-DoF dexterous joint action space, and we compare against two alternatives:
(i) a \emph{wrist-only} representation that removes all finger-level supervision, and
(ii) a \emph{fingertip-based} representation~\cite{yang2025egovla} that predicts $\mathbb{SE}(3)$ trajectories of the wrist and fingertips, followed by an MLP mapping to robot joint commands.\looseness=-1

As shown in Figure~\ref{fig:action_space_ablation}, the action representation strongly influences task performance.
The wrist-only representation performs poorly across all tasks, particularly those requiring precise finger articulation and contact timing (e.g., \emph{Tongs}, \emph{Cards}).
In these cases, policies frequently grasp tools too weakly or at unstable heights, or close the hand prematurely, resulting in missed or brittle contacts.

The fingertip-based representation provides richer geometric supervision and improves performance on some tasks, but remains inconsistent.
Small errors in fingertip pose often lead to implausible joint configurations after mapping, causing unstable grasps or contact loss in contact-sensitive tasks such as \emph{Cards} and \emph{Bottle}.
In contrast, pretraining with retargeted joint-space hand actions yields the most consistent performance across all tasks.
We therefore use retargeted joint-space hand actions as a practical and effective choice for large-scale human pretraining.\looseness=-1

\section{Related Work}

\textbf{Robot Learning from Human Data.}
Human demonstrations have been widely used to scale robot learning, with early works leveraging human videos primarily for representation learning or intent inference \cite{nair2022r3m,xiao2022masked,majumdar2023we,lynch2020learning, zheng2025flare}. 
Subsequent approaches use human data to guide planning or high-level control while relying on robot demonstrations for low-level execution \cite{wang2023mimicplay, young2021visual, zheng2024tracevla, niu2024llarva,xu2023xskill,xu2024flow}. 
More recent methods exploit advances in egocentric sensing and 3D hand tracking to treat human videos as dense action supervision. EgoMimic~\cite{kareer2024egomimicscalingimitationlearning}, ~\citet{qiu2025humanoid}, and DexWild~\cite{tao2025dexwild} co-train a unified imitation policy on human and robot demonstrations through explicit alignment, while EgoVLA~\cite{yang2025egovla} pretrains a VLA model on human hand motion and transfers it to robots via inverse kinematics and retargeting. 
Concurrent work~\cite{kareer2025emergencehumanrobottransfer} demonstrates that VLA pretrained on large-scale diverse cross-embodiment data can unlock human to robot transfer capability.

In contrast to these works, our work focuses on \emph{pure large-scale human pretraining} with minimal robot supervision, with the goal of learning both wrist motion and dexterous hand articulation directly from diverse egocentric human videos. Compared to approaches that primarily transfer wrist motion for gripper-based platforms, our setting retains rich hand motion information that is critical for dexterous manipulation. Compared to recent methods that also leverage hand-level supervision, we pretrain on substantially larger-scale human data and systematically show that scaling human video yields consistent improvements in both human validation metrics and downstream robot performance, enabling strong dexterous manipulation performance on advanced multi-DoF robotic hands. \looseness=-1

\textbf{Scaling Properties in Robot Learning.} Inspired by scaling laws observed in language and vision, recent work has begun to investigate whether similar principles govern robot learning. Empirically, large-scale robot datasets and foundation-style policies show that increasing data diversity and coverage leads to improved robustness and generalization across tasks and environments~\cite{kalashnikov2018scalable,zitkovich2023rt,open_x_embodiment_rt_x_2023,walke2023bridgedata,team2024octo, brohan2022rt}. \citet{hu2024data} shows that policy generalization follows an approximate power law with environment and object diversity, while additional demonstrations quickly saturate, highlighting that diversity is more important than raw data volume. This is consistent with prior work emphasizing efficient data collection through compositional diversity~\cite{gao2024efficient,xie2024decomposing}. Compared to prior work that primarily scales robot-collected data, we show that scaling diverse in-the-wild human egocentric data leads to systematic gains in dexterous manipulation, establishing human video as an efficient and scalable supervision source.

\textbf{Learning Dexterous Manipulation.}
Dexterous manipulation has evolved from analytic and control-based grasping methods that model force closure, contact stability, and hand kinematics \cite{ponce1993characterizing,ponce1997computing,rodriguez2012caging,rosales2012synthesis,prattichizzo2012manipulability,dai2017synthesis} to learning-based approaches that acquire contact-rich behaviors from data \cite{andrychowicz2020learning,nagabandi2020deep,xu2025dexumiusinghumanhand}. Subsequent works introduce structured representations such as grasp affordances, contact maps, and hand–object interaction fields to better capture dexterous geometry and physics \cite{brahmbhatt2019contactgrasp,corona2020ganhand,jiang2021hand,yang2021cpf,turpin2022grasp}. More recent methods aim to learn generalizable multi-fingered manipulation policies with unified perception and control~\cite{shao2020unigrasp,wu2022learning}. However, scaling dexterous manipulation remains challenging due to the high-dimensional action space, the cost of robot data collection, and the limited capabilities of current dexterous hand hardware. Our work leverages egocentric human videos with dense hand tracking as an alternative supervision source, showing that human hand motion provides transferable signals for improving robotic dexterous manipulation.

\section{Acknowledgement}
This work would not have been possible without the dedication and expertise of our robot operators: Ivy Tam, Ashley Kim, Aly Khater, Rhea Alve, Noah Huang, Ty Seligman, Mahnoor Kareem, Christian Soto, Chaitanya Kothapalli, Kendra Shu, Rex Asato, Eley Barba Preciados, and April Zitkovich, who provided invaluable support in large-scale data collection and evaluation.
We are also grateful to Johan Bjorck, Scott Reed, Runyu Ding and Joel Jang for their insightful discussions and feedback throughout the project.

\section{Conclusion}
In this work, we show that effective human-to-robot transfer for dexterous manipulation is fundamentally a scaling phenomenon.
By pretraining a vision–language–action policy on over 20K hours of egocentric human manipulation data, \ours~uncovers a clear log-linear scaling law between human action prediction loss and data scale, and demonstrates that this loss strongly predicts downstream real-robot performance.
Beyond scale, we identify a simple and effective transfer recipe: combining large-scale human pretraining with a small amount of aligned human–robot mid-training enables strong long-horizon dexterous manipulation, emergent one-shot adaptation, and robust transfer across robot embodiments with substantially different kinematics and hand designs.

Looking forward, several directions remain open.
While we observe no saturation within the explored regime, jointly scaling human data and model capacity may unlock further gains, including improved long-horizon planning and compositional generalization.
As egocentric human data continues to grow, incorporating weaker or unlabeled video via self-supervised objectives may further amplify these benefits.
Finally, as robotic hardware becomes more human-like in kinematics and dexterity, the embodiment gap will naturally shrink, enabling stronger transfer and potentially zero-shot execution on novel tasks.
Together, these results point toward a future where humans can be treated as a truly scalable embodiment for learning general embodied intelligence.

\clearpage
\appendix

\section{Details of Robot System}
\begin{figure}[!htbp]
    \centering
    \includegraphics[width=\linewidth]
    {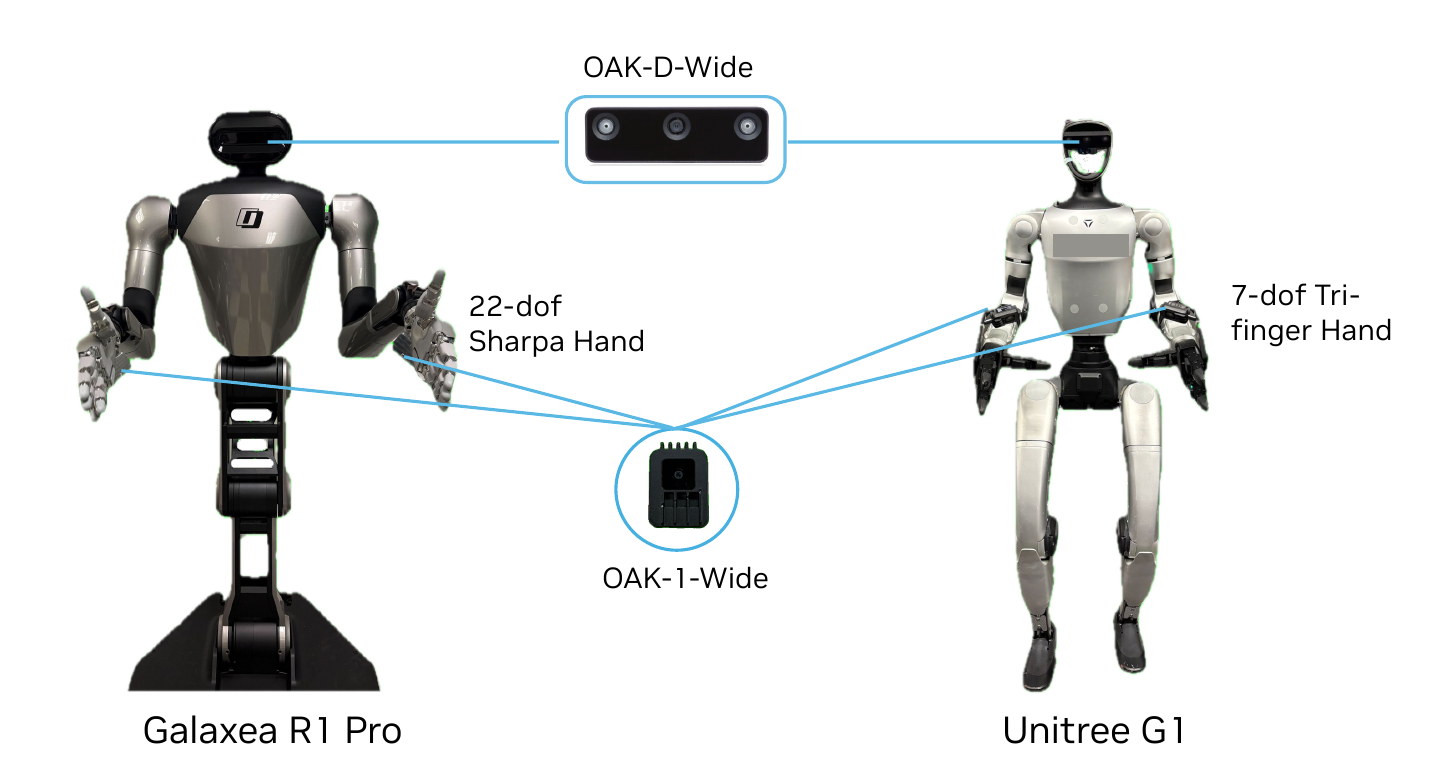}
    \caption{Visualization of the robot system setups on the Galaxea R1 Pro and Unitree G1 platforms.
The Galaxea R1 Pro is equipped with 22-DoF Sharpa dexterous hands, while the Unitree G1 is mounted with 7-DoF tri-finger hands. Both robots have with two OAK-1-Wide cameras for the wrist view and an OAK-D-Wide head-mounted camera to provide egocentric visual observations.
    }
\label{fig:robot_system}
    \vspace{-0.5em}
\end{figure}
\section{Task Descriptions and Evaluation Rubric}
\label{sec:task_rubric}

In this section, we provide detailed descriptions of the post-training evaluation tasks introduced in Section~\ref{sec:exp}, together with the corresponding evaluation rubrics.
Each task is specified by a natural-language instruction and is evaluated using a task completion score in $[0,1]$, designed to capture partial progress in addition to binary success.

Depending on the structure of the task, we adopt one of two scoring strategies.
For tasks that naturally decompose into a sequence of independent, well-defined sub-skills (e.g., tool grasping, placement, or discrete manipulation steps), we use an \emph{additive rubric}, where the final score is the sum of achieved sub-scores.
For tasks involving deformable objects or tightly coupled manipulation stages, where intermediate states are difficult to define independently, we instead use a \emph{progress-based rubric}, where the score reflects the furthest task milestone achieved.
Both designs aim to provide fine-grained, interpretable feedback while remaining faithful to the underlying task structure.
For each task, unless specifically mentioned, we evaluate each checkpoint for 10 trials.

\subsection*{R1 Pro Sharpa}

\begin{itemize}

\item \textbf{Task I: Shirt Rolling.} 
\textit{Text Instruction:} \emph{``Roll the T-shirt and put it into the basket.''}  
The robot must coordinate both hands to progressively fold, roll, and place a deformable T-shirt into a basket.

\textit{Grading rubric (progress-based):}
\begin{itemize}
    \item $0.0$: No folding behavior.
    \item $0.3$: Executes an initial fold.
    \item $0.5$: Executes multiple folds or partial rolling.
    \item $0.8$: Completes a continuous rolling motion into a compact shape.
    \item $1.0$: Places the rolled shirt into the basket.
\end{itemize}

Evaluation trials: 10 trials.

\item \textbf{Task II: Card Sorting.}  
\textit{Text Instruction:} \emph{``Pick up the card and sort it into the correct card holder.''}  
The robot must separate a single card from a tightly stacked deck and insert it into the correct holder.

\textit{Grading rubric (progress-based):}
\begin{itemize}
    \item $0.0$: No successful pickup; entire deck disturbed or lifted.
    \item $0.3$: Poor grasp; multiple cards lifted unintentionally.
    \item $0.5$: Single card successfully picked up.
    \item $0.7$: Card placed with noticeable disturbance or incorrect insertion.
    \item $0.9$: Correct card placed correctly with minor disturbance.
    \item $1.0$: Single correct card cleanly placed into the correct holder.
\end{itemize}

\item \textbf{Task III: Tong Fruit Transfer.}  
\textit{Text Instruction:} \emph{``Use the tong to pick up the fruit and place it into the basket.''}  
The robot must first grasp the tong and then use the tool to pick up a target object and place it into a container.
This task is evaluated with 2 random fruits, lemon and plum, each with 5 evaluation trials.

\textit{Grading rubric (additive):}
\begin{itemize}
    \item ${+0.4}$: Successfully grasps the tongs.
    \item ${+0.2}$: Picks up the fruit using the tongs.
    \item ${+0.2}$: Places the fruit into the basket.
    \item ${+0.2}$: Returns the tongs to the table.
\end{itemize}

\item \textbf{Task IV: Bottle Cap Unscrewing.}  
\textit{Text Instruction:} \emph{``Unscrew the cap from the bottle.''}  
The robot must grasp a bottle, rotate the cap multiple times, remove it, and place it on the table.
To test generalization across object variations, we evaluate 4 bottle configurations with different heights and cap sizes. 
Performance is measured over 12 trials in total, with 4 trials conducted for each bottle.

\textit{Grading rubric (additive):}
\begin{itemize}
    \item $+0.1$: Grasps the bottle.
    \item $+0.5$: Unscrews the cap with at least three continuous rotations.
    \item $+0.2$: Fully removes the cap.
    \item $+0.2$: Places the cap on the table.
\end{itemize}

\item \textbf{Task V: Syringe Liquid Transfer.}  
\textit{Text Instruction:} \emph{``Pick up the syringe, draw liquid from tube A, inject it into tube B, and throw the syringe into the trash can.''}  
This task requires precise tool manipulation, spatial alignment, and long-horizon sequencing.

\textit{Grading rubric (additive):}
\begin{itemize}
    \item $+0.1$: Picks up the syringe.
    \item $+0.1$: Correctly aims at tube A.
    \item $+0.2$: Pulls the plunger to draw liquid.
    \item $+0.1$: Re-aims at tube B.
    \item $+0.2$: Pushes the plunger to inject liquid.
    \item $+0.2$: Hands over or releases the syringe.
    \item $+0.1$: Disposes of the syringe into the trash can.
\end{itemize}

To evaluate one-shot adaptation, we consider two additional manipulation tasks on the R1 Pro Sharpa platform. 
For each task, the policy is post-trained using a single robot demonstration, and performance is evaluated using a task-specific completion score in $[0,1]$.

\item \textbf{Task VIII: One-Shot T-Shirt Folding.}  
\textit{Text Instruction:} \emph{``Fold the T-shirt.''}  

\textit{Grading rubric (additive):}
\begin{itemize}
    \item $+0.4$: Successfully folds at least one sleeve.
    \item $+0.4$: Successfully folds both sleeves.
    \item $+0.3$: Executes a bottom fold.
\end{itemize}
During the evaluation, we also subtract a 0.1 if it does one of the fold but the folding is messy.

\item \textbf{Task IX: One-Shot Bottle Cap Unscrewing.}  
\textit{Text Instruction:} \emph{``Unscrew the cap from the water bottle.''}  
The task decomposes naturally into discrete, well-defined manipulation steps and is evaluated using additive scoring.

\textit{Grading rubric (additive):}
\begin{itemize}
    \item $+0.1$: Grasps the bottle.
    \item $+0.2$: Performs one to two successful unscrewing rotations.
    \item $+0.5$: Performs at least three continuous unscrewing rotations.
    \item $+0.2$: Fully removes the cap.
    \item $+0.2$: Places the cap on the table.

\end{itemize}

\end{itemize}

For all tasks, we additionally report a binary task success rate, where a trial is considered successful only if the task is completed end-to-end according to the instruction.

\subsection*{G1 Tasks}

In addition to the R1 Pro Sharpa evaluations, we evaluate cross-embodiment generalization on the Unitree G1 platform using two manipulation tasks. 
Both tasks are evaluated using additive scoring, reflecting their decomposition into discrete, well-defined sub-skills.

\begin{itemize}

\item \textbf{Pen in Bin.}  
\textit{Text Instruction:} \emph{``Marker canister task.''}  
The robot must open a canister, pick up a marker, and place the marker into the canister.

\textit{Grading rubric (additive):}
\begin{itemize}
    \item $+0.25$: Picks up or opens the canister.
    \item $+0.25$: Places the canister down stably.
    \item $+0.25$: Picks up the marker.
    \item $+0.25$: Places the marker into the canister.
\end{itemize}

\item \textbf{Dish Handover in Rack}  
\textit{Text Instruction:} \emph{``Put plates on dishrack.''}  
The robot is presented with three plates on a table and must transfer them, potentially between hands, and place them upright into a dish rack.

\textit{Grading rubric (additive, per plate):}
For each plate, we assign:
\begin{itemize}
    \item $0.11$: Picks up the plate.
    \item $0.11$: Transfers the plate between hands.
    \item $0.11$: Places the plate upright into the dish rack.
\end{itemize}

The final task completion score is computed by summing scores across all three plates, yielding a maximum score of $1.0$ per trial.
\end{itemize}

\section{Details of In-the-wild Human Data Curation}
\begin{figure*}[!thbp]
    \vspace{-0.5em}
    \centering
    \includegraphics[width=0.95\linewidth]{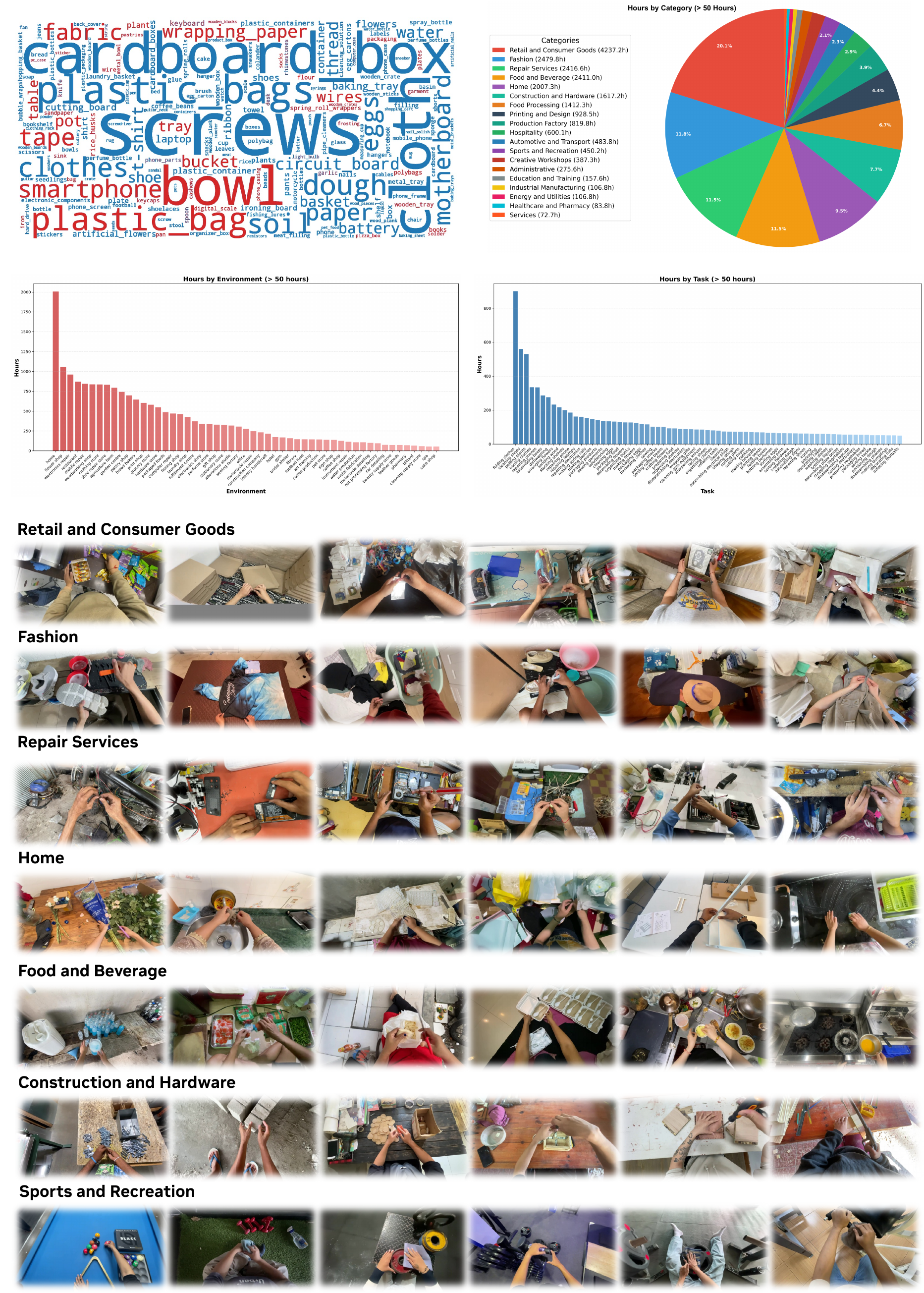}
    \caption{
Statistical distributions and qualitative examples of the egocentric human activity dataset, showing long-tailed coverage across categories, environments, tasks, and objects.
}
\label{fig:dataset_overview}
\vspace{-0.5em}
\end{figure*}

\subsection{Dataset Statistics and Analysis}
\label{sec:dataset_stats}

This section provides a detailed analysis of the large-scale egocentric human activity data used for Stage~I pretraining. Figure~\ref{fig:dataset_overview} summarizes the dataset from multiple complementary perspectives, including statistical distributions shown in the top row and qualitative egocentric examples shown in the bottom row. Specifically, the upper-left panel visualizes the object vocabulary, the upper-middle panel shows the category distribution, the upper-right panel presents the distribution of environments, and the rightmost panel in the second row illustrates the task distribution. Together, these statistics reveal long-tailed distributions across categories, environments, tasks, and objects, highlighting both the scale and diversity of real-world manipulation behaviors captured in the dataset.

\subsection{Category Distribution}
The category-level distribution of total video hours (only more than 50 hours categories are counted in the plot) is shown in the upper-middle panel of Fig.~\ref{fig:dataset_overview}. The dataset is dominated by retail and consumer goods (20.1\%), followed by fashion (11.8\%), repair services (11.5\%), and food and beverage (11.5\%). Additional substantial portions come from home environments (9.5\%), construction and hardware (7.7\%), food processing (6.7\%), and printing and design (4.4\%), while the remaining categories form a long tail spanning hospitality, automotive and transport, sports and recreation, education, healthcare, energy, and industrial domains.

\subsection{Environment Diversity}
The distribution of data across environments with more than 50 hours of recordings is shown in the lower-left bar chart of Fig.~\ref{fig:dataset_overview}. The most frequent environments include homes, flower shops, electronics repair shops, furniture repair shops, woodworking shops, and clothing stores. Beyond these dominant settings, dozens of additional environments—such as grocery stores, bakeries, workshops, factories, studios, libraries, and other service-oriented spaces—contribute smaller but non-negligible amounts of data, resulting in a long-tailed environment distribution.

\subsection{Task Coverage}
Task-level statistics are presented in the lower-right bar chart of Fig.~\ref{fig:dataset_overview}. High-frequency tasks include folding clothes, cleaning shoes, potting plants, ironing, assembling boxes, arranging flowers, sanding wood, and food preparation. In addition, a wide range of medium- and low-frequency tasks cover packing, sorting, repairing, decorating, cutting, bottling, inflating, and disassembling. Many tasks involve multi-step procedures and sustained physical interaction, reflecting realistic, long-horizon manipulation behaviors rather than isolated pick-and-place actions.

\subsection{Object Vocabulary}
The object vocabulary is visualized using a word cloud in the upper-left panel of Fig.~\ref{fig:dataset_overview}. Frequently occurring objects include cardboard boxes, plastic bags, bowls, trays, tape, clothes, dough, batteries, and circuit boards, alongside a long tail of less frequent objects. The diversity of object instances and their co-occurrence with a wide range of tasks indicate strong compositional structure, supporting the learning of transferable action representations across objects, tasks, and environments.


\section{Hand Retargeting Algorithm}

We retarget 21 human hand keypoints (represented as 25 keypoints per hand with 3D position and orientation) into the 22-DoF joint space of the Sharpa Hand~\citep{sharpa_wave} using a per-frame, optimization-based procedure. 
The robot hand is modeled using URDF-based forward kinematics, which maps joint angles to 20 robot keypoint poses (positions and quaternions).

For each hand and each timestep, we solve a nonlinear program over the 22 joint angles, subject only to joint limits from the URDF, and minimize a weighted combination of different objectives. The optimization is implemented in CasADi and solved using IPOPT, warm-started from the previous frame’s solution. 
The resulting joint angles are further smoothed using a first-order exponential filter, to reduce temporal jitter.
This design enforces joint limits and kinematic consistency while preserving finger articulation and pinch/fist semantics, and yields the 22-DoF action space used for pretraining and for interfacing with the target robot.
\looseness=-1










\subsection{Additional Details on Cross-Embodiment Transfer to G1}
\label{app:g1_transfer}

This subsection provides additional implementation details on how we adapt the human-pretrained policy to the Unitree G1 robot, which differs substantially from the primary Galaxea R1 Pro platform in both kinematics and hand actuation.

\paragraph{Shared Wrist Action.}
Across all embodiments, we represent arm motion using relative end-effector poses in $SE(3)$, defined by frame-to-frame wrist transformations.
This representation is shared between human demonstrations and robot executions, and remains invariant to differences in absolute workspace, camera placement, and arm kinematics.

\paragraph{Hand Action Adaptation.}
Human pretraining is performed in a 22-DoF dexterous hand joint space obtained via retargeted human hand motion.
To support robots with different embodiments, such as the Unitree G1 with a 7-DoF tri-finger hand, we introduce lightweight embodiment-conditioned MLP adapters at both the input and output interfaces of the DiT action module, following the design of GR00T-N1 and N1.5~\citep{nvidia2025gr00tn1openfoundation}.
Specifically, embodiment-specific encoders map robot proprioceptive state and noisy action inputs into a shared latent action space, while embodiment-specific decoders map the DiT outputs back to the corresponding joint action space.
The vision–language backbone and the DiT action expert are fully shared across all embodiments. \looseness=-1

\paragraph{Embodiment-Specific Mid-Training.}
To ground the human-pretrained representations in the G1 control space, we include the G1 robot play data during the aligned mid-training stage.
During this stage, only the vision encoder, DiT action expert, and state-action encoder and decoder are updated, while the vision-language backbone remains frozen.
This design allows the model to preserve human-derived manipulation structure while adapting to the G1’s sensing and actuation interfaces.

\paragraph{Discussion.}
Importantly, the G1 is never trained from scratch.
Instead, mid-training serves to align an already learned, human-derived manipulation representation with a new embodiment.
As shown in Section~\ref{sec:exp}, this approach yields substantially higher performance than training directly on G1 data alone, indicating that large-scale human pretraining provides a reusable and embodiment-agnostic motor prior that can be efficiently adapted to robots with different kinematics and hand designs.

\clearpage
\setcitestyle{numbers}
\bibliographystyle{plainnat}
\bibliography{main}

@misc{hoque2025egodex,
      title={EgoDex: Learning Dexterous Manipulation from Large-Scale Egocentric Video}, 
      author={Ryan Hoque and Peide Huang and David J. Yoon and Mouli Sivapurapu and Jian Zhang},
      year={2025},
      eprint={2505.11709},
      archivePrefix={arXiv},
      primaryClass={cs.CV},
      url={https://arxiv.org/abs/2505.11709}, 
}

@inproceedings{
qiu2025humanoid,
title={Humanoid Policy {\textasciitilde} Human Policy},
author={Ri-Zhao Qiu and Shiqi Yang and Xuxin Cheng and Chaitanya Chawla and Jialong Li and Tairan He and Ge Yan and David J. Yoon and Ryan Hoque and Lars Paulsen and Ge Yang and Jian Zhang and Sha Yi and Guanya Shi and Xiaolong Wang},
booktitle={9th Annual Conference on Robot Learning},
year={2025},
url={https://openreview.net/forum?id=Tx54fkQ3Cq}
}

@misc{kareer2024egomimicscalingimitationlearning,
      title={EgoMimic: Scaling Imitation Learning via Egocentric Video}, 
      author={Simar Kareer and Dhruv Patel and Ryan Punamiya and Pranay Mathur and Shuo Cheng and Chen Wang and Judy Hoffman and Danfei Xu},
      year={2024},
      eprint={2410.24221},
      archivePrefix={arXiv},
      primaryClass={cs.RO},
      url={https://arxiv.org/abs/2410.24221}, 
}

@misc{nvidia2025gr00tn1openfoundation,
      title={GR00T N1: An Open Foundation Model for Generalist Humanoid Robots}, 
      author={NVIDIA and : and Johan Bjorck and Fernando Castañeda and Nikita Cherniadev and Xingye Da and Runyu Ding and Linxi "Jim" Fan and Yu Fang and Dieter Fox and Fengyuan Hu and Spencer Huang and Joel Jang and Zhenyu Jiang and Jan Kautz and Kaushil Kundalia and Lawrence Lao and Zhiqi Li and Zongyu Lin and Kevin Lin and Guilin Liu and Edith Llontop and Loic Magne and Ajay Mandlekar and Avnish Narayan and Soroush Nasiriany and Scott Reed and You Liang Tan and Guanzhi Wang and Zu Wang and Jing Wang and Qi Wang and Jiannan Xiang and Yuqi Xie and Yinzhen Xu and Zhenjia Xu and Seonghyeon Ye and Zhiding Yu and Ao Zhang and Hao Zhang and Yizhou Zhao and Ruijie Zheng and Yuke Zhu},
      year={2025},
      eprint={2503.14734},
      archivePrefix={arXiv},
      primaryClass={cs.RO},
      url={https://arxiv.org/abs/2503.14734}, 
}

@article{wang2023mimicplay,
    title={Mimicplay: Long-horizon imitation learning by watching human play},
    author={Wang, Chen and Fan, Linxi and Sun, Jiankai and Zhang, Ruohan and Fei-Fei, Li and Xu, Danfei and Zhu, Yuke and
    Anandkumar, Anima},
    journal={arXiv preprint arXiv:2302.12422},
    year={2023}
}

@inproceedings{xu2023xskill,
  title={Xskill: Cross embodiment skill discovery},
  author={Xu, Mengda and Xu, Zhenjia and Chi, Cheng and Veloso, Manuela and Song, Shuran},
  booktitle={Conference on robot learning},
  pages={3536--3555},
  year={2023},
  organization={PMLR}
}

@inproceedings{punamiyaegobridge,
  title={EgoBridge: Domain Adaptation for Generalizable Imitation from Egocentric Human Data},
  author={Punamiya, Ryan and Patel, Dhruv and Aphiwetsa, Patcharapong and Kuppili, Pranav and Zhu, Lawrence Y and Kareer, Simar and Hoffman, Judy and Xu, Danfei},
  booktitle={The Thirty-ninth Annual Conference on Neural Information Processing Systems}
}

@misc{open_x_embodiment_rt_x_2023,
title={Open {X-E}mbodiment: Robotic Learning Datasets and {RT-X} Models},
author = {{Open X-Embodiment Collaboration} and others},
howpublished  = {International Conference on Robotics and Automation},
year = {2024},
}

@article{brohan2022rt,
  title={Rt-1: Robotics transformer for real-world control at scale},
  author={Brohan, Anthony and Brown, Noah and Carbajal, Justice and Chebotar, Yevgen and Dabis, Joseph and Finn, Chelsea and Gopalakrishnan, Keerthana and Hausman, Karol and Herzog, Alex and Hsu, Jasmine and others},
  journal={arXiv preprint arXiv:2212.06817},
  year={2022}
}

@article{nair2022r3m,
  title={R3m: A universal visual representation for robot manipulation},
  author={Nair, Suraj and Rajeswaran, Aravind and Kumar, Vikash and Finn, Chelsea and Gupta, Abhinav},
  journal={arXiv preprint arXiv:2203.12601},
  year={2022}
}

@inproceedings{walke2023bridgedata,
    title={BridgeData V2: A Dataset for Robot Learning at Scale},
    author={Walke, Homer and Black, Kevin and Lee, Abraham and Kim, Moo Jin and Du, Max and Zheng, Chongyi and Zhao, Tony and Hansen-Estruch, Philippe and Vuong, Quan and He, Andre and Myers, Vivek and Fang, Kuan and Finn, Chelsea and Levine, Sergey},
    booktitle={Conference on Robot Learning (CoRL)},
    year={2023}
}

@misc{sharpa_wave,
  title        = {Sharpa Wave},
  author       = {{Sharpa Robotics}},
  howpublished = {\url{https://www.sharpa.com/pages/wave}},
  year         = {2024},
  note         = {Accessed: 2026-01-31}
}

@misc{tao2025dexwild,
      title={DexWild: Dexterous Human Interactions for In-the-Wild Robot Policies}, 
      author={Tony Tao and Mohan Kumar Srirama and Jason Jingzhou Liu and Kenneth Shaw and Deepak Pathak},
      year={2025},
      eprint={2505.07813},
      archivePrefix={arXiv},
      primaryClass={cs.RO},
      url={https://arxiv.org/abs/2505.07813}, 
}

@inproceedings{
zheng2025flare,
title={{FLARE}: Robot Learning with Implicit World Modeling},
author={Ruijie Zheng and Jing Wang and Scott Reed and Johan Bjorck and Yu Fang and Fengyuan Hu and Joel Jang and Kaushil Kundalia and Zongyu Lin and Lo{\"\i}c Magne and Avnish Narayan and You Liang Tan and Guanzhi Wang and Qi Wang and Jiannan Xiang and Yinzhen Xu and Seonghyeon Ye and Jan Kautz and Furong Huang and Yuke Zhu and Linxi Fan},
booktitle={9th Annual Conference on Robot Learning},
year={2025},
url={https://openreview.net/forum?id=HXJ6pUSn1L}
}

@misc{kareer2025emergencehumanrobottransfer,
      title={Emergence of Human to Robot Transfer in Vision-Language-Action Models}, 
      author={Simar Kareer and Karl Pertsch and James Darpinian and Judy Hoffman and Danfei Xu and Sergey Levine and Chelsea Finn and Suraj Nair},
      year={2025},
      eprint={2512.22414},
      archivePrefix={arXiv},
      primaryClass={cs.RO},
      url={https://arxiv.org/abs/2512.22414}, 
}

@misc{ben2025homiehumanoidlocomanipulationisomorphic,
      title={HOMIE: Humanoid Loco-Manipulation with Isomorphic Exoskeleton Cockpit}, 
      author={Qingwei Ben and Feiyu Jia and Jia Zeng and Junting Dong and Dahua Lin and Jiangmiao Pang},
      year={2025},
      eprint={2502.13013},
      archivePrefix={arXiv},
      primaryClass={cs.RO},
      url={https://arxiv.org/abs/2502.13013}, 
}

@article{xiao2022masked,
  title={Masked visual pre-training for motor control},
  author={Xiao, Tete and Radosavovic, Ilija and Darrell, Trevor and Malik, Jitendra},
  journal={arXiv preprint arXiv:2203.06173},
  year={2022}
}

@article{majumdar2023we,
  title={Where are we in the search for an artificial visual cortex for embodied intelligence?},
  author={Majumdar, Arjun and Yadav, Karmesh and Arnaud, Sergio and Ma, Jason and Chen, Claire and Silwal, Sneha and Jain, Aryan and Berges, Vincent-Pierre and Wu, Tingfan and Vakil, Jay and others},
  journal={Advances in Neural Information Processing Systems},
  volume={36},
  pages={655--677},
  year={2023}
}

@inproceedings{lynch2020learning,
  title={Learning latent plans from play},
  author={Lynch, Corey and Khansari, Mohi and Xiao, Ted and Kumar, Vikash and Tompson, Jonathan and Levine, Sergey and Sermanet, Pierre},
  booktitle={Conference on robot learning},
  pages={1113--1132},
  year={2020},
  organization={Pmlr}
}

@article{yang2025egovla,
  title={Egovla: Learning vision-language-action models from egocentric human videos},
  author={Yang, Ruihan and Yu, Qinxi and Wu, Yecheng and Yan, Rui and Li, Borui and Cheng, An-Chieh and Zou, Xueyan and Fang, Yunhao and Cheng, Xuxin and Qiu, Ri-Zhao and others},
  journal={arXiv preprint arXiv:2507.12440},
  year={2025}
}

@inproceedings{young2021visual,
  title={Visual imitation made easy},
  author={Young, Sarah and Gandhi, Dhiraj and Tulsiani, Shubham and Gupta, Abhinav and Abbeel, Pieter and Pinto, Lerrel},
  booktitle={Conference on Robot learning},
  pages={1992--2005},
  year={2021},
  organization={PMLR}
}

@article{zheng2024tracevla,
  title={Tracevla: Visual trace prompting enhances spatial-temporal awareness for generalist robotic policies},
  author={Zheng, Ruijie and Liang, Yongyuan and Huang, Shuaiyi and Gao, Jianfeng and Daum{\'e} III, Hal and Kolobov, Andrey and Huang, Furong and Yang, Jianwei},
  journal={arXiv preprint arXiv:2412.10345},
  year={2024}
}

@article{niu2024llarva,
  title={Llarva: Vision-action instruction tuning enhances robot learning},
  author={Niu, Dantong and Sharma, Yuvan and Biamby, Giscard and Quenum, Jerome and Bai, Yutong and Shi, Baifeng and Darrell, Trevor and Herzig, Roei},
  journal={arXiv preprint arXiv:2406.11815},
  year={2024}
}

@inproceedings{ponce1993characterizing,
  title={On characterizing and computing three-and four-finger force-closure grasps of polyhedral objects},
  author={Ponce, Jean and Sullivan, Steve and Boissonnat, J-D and Merlet, J-P},
  booktitle={[1993] Proceedings IEEE International Conference on Robotics and Automation},
  pages={821--827},
  year={1993},
  organization={IEEE}
}

@article{ponce1997computing,
  title={On computing four-finger equilibrium and force-closure grasps of polyhedral objects},
  author={Ponce, Jean and Sullivan, Steve and Sudsang, Attawith and Boissonnat, Jean-Daniel and Merlet, Jean-Pierre},
  journal={The International Journal of Robotics Research},
  volume={16},
  number={1},
  pages={11--35},
  year={1997},
  publisher={Sage Publications Sage CA: Thousand Oaks, CA}
}

@article{rodriguez2012caging,
  title={From caging to grasping},
  author={Rodriguez, Alberto and Mason, Matthew T and Ferry, Steve},
  journal={The International Journal of Robotics Research},
  volume={31},
  number={7},
  pages={886--900},
  year={2012},
  publisher={SAGE Publications Sage UK: London, England}
}

@inproceedings{rosales2012synthesis,
  title={On the synthesis of feasible and prehensile robotic grasps},
  author={Rosales, Carlos and Su{\'a}rez, Ra{\'u}l and Gabiccini, Marco and Bicchi, Antonio},
  booktitle={2012 IEEE international conference on robotics and automation},
  pages={550--556},
  year={2012},
  organization={IEEE}
}

@article{prattichizzo2012manipulability,
  title={On the manipulability ellipsoids of underactuated robotic hands with compliance},
  author={Prattichizzo, Domenico and Malvezzi, Monica and Gabiccini, Marco and Bicchi, Antonio},
  journal={Robotics and Autonomous Systems},
  volume={60},
  number={3},
  pages={337--346},
  year={2012},
  publisher={Elsevier}
}

@incollection{dai2017synthesis,
  title={Synthesis and optimization of force closure grasps via sequential semidefinite programming},
  author={Dai, Hongkai and Majumdar, Anirudha and Tedrake, Russ},
  booktitle={Robotics Research: Volume 1},
  pages={285--305},
  year={2017},
  publisher={Springer}
}

@article{andrychowicz2020learning,
  title={Learning dexterous in-hand manipulation},
  author={Andrychowicz, OpenAI: Marcin and Baker, Bowen and Chociej, Maciek and Jozefowicz, Rafal and McGrew, Bob and Pachocki, Jakub and Petron, Arthur and Plappert, Matthias and Powell, Glenn and Ray, Alex and others},
  journal={The International Journal of Robotics Research},
  volume={39},
  number={1},
  pages={3--20},
  year={2020},
  publisher={SAGE Publications Sage UK: London, England}
}

@inproceedings{nagabandi2020deep,
  title={Deep dynamics models for learning dexterous manipulation},
  author={Nagabandi, Anusha and Konolige, Kurt and Levine, Sergey and Kumar, Vikash},
  booktitle={Conference on robot learning},
  pages={1101--1112},
  year={2020},
  organization={PMLR}
}

@inproceedings{brahmbhatt2019contactgrasp,
  title={Contactgrasp: Functional multi-finger grasp synthesis from contact. In 2019 IEEE},
  author={Brahmbhatt, Samarth and Handa, Ankur and Hays, James and Fox, Dieter},
  booktitle={RSJ International Conference on Intelligent Robots and Systems (IROS)},
  pages={2386--2393},
  year={2019}
}

@inproceedings{corona2020ganhand,
  title={Ganhand: Predicting human grasp affordances in multi-object scenes},
  author={Corona, Enric and Pumarola, Albert and Alenya, Guillem and Moreno-Noguer, Francesc and Rogez, Gr{\'e}gory},
  booktitle={Proceedings of the IEEE/CVF conference on computer vision and pattern recognition},
  pages={5031--5041},
  year={2020}
}

@inproceedings{jiang2021hand,
  title={Hand-object contact consistency reasoning for human grasps generation},
  author={Jiang, Hanwen and Liu, Shaowei and Wang, Jiashun and Wang, Xiaolong},
  booktitle={Proceedings of the IEEE/CVF international conference on computer vision},
  pages={11107--11116},
  year={2021}
}

@inproceedings{yang2021cpf,
  title={Cpf: Learning a contact potential field to model the hand-object interaction},
  author={Yang, Lixin and Zhan, Xinyu and Li, Kailin and Xu, Wenqiang and Li, Jiefeng and Lu, Cewu},
  booktitle={Proceedings of the IEEE/CVF international conference on computer vision},
  pages={11097--11106},
  year={2021}
}

@inproceedings{turpin2022grasp,
  title={Grasp’d: Differentiable contact-rich grasp synthesis for multi-fingered hands},
  author={Turpin, Dylan and Wang, Liquan and Heiden, Eric and Chen, Yun-Chun and Macklin, Miles and Tsogkas, Stavros and Dickinson, Sven and Garg, Animesh},
  booktitle={European Conference on Computer Vision},
  pages={201--221},
  year={2022},
  organization={Springer}
}

@article{shao2020unigrasp,
  title={Unigrasp: Learning a unified model to grasp with multifingered robotic hands},
  author={Shao, Lin and Ferreira, Fabio and Jorda, Mikael and Nambiar, Varun and Luo, Jianlan and Solowjow, Eugen and Ojea, Juan Aparicio and Khatib, Oussama and Bohg, Jeannette},
  journal={IEEE Robotics and Automation Letters},
  volume={5},
  number={2},
  pages={2286--2293},
  year={2020},
  publisher={IEEE}
}

@article{wu2022learning,
  title={Learning diverse and physically feasible dexterous grasps with generative model and bilevel optimization},
  author={Wu, Albert and Guo, Michelle and Liu, C Karen},
  journal={arXiv preprint arXiv:2207.00195},
  year={2022}
}

@inproceedings{kalashnikov2018scalable,
  title={Scalable deep reinforcement learning for vision-based robotic manipulation},
  author={Kalashnikov, Dmitry and Irpan, Alex and Pastor, Peter and Ibarz, Julian and Herzog, Alexander and Jang, Eric and Quillen, Deirdre and Holly, Ethan and Kalakrishnan, Mrinal and Vanhoucke, Vincent and others},
  booktitle={Conference on robot learning},
  pages={651--673},
  year={2018},
  organization={PMLR}
}

@inproceedings{zitkovich2023rt,
  title={Rt-2: Vision-language-action models transfer web knowledge to robotic control},
  author={Zitkovich, Brianna and Yu, Tianhe and Xu, Sichun and Xu, Peng and Xiao, Ted and Xia, Fei and Wu, Jialin and Wohlhart, Paul and Welker, Stefan and Wahid, Ayzaan and others},
  booktitle={Conference on Robot Learning},
  pages={2165--2183},
  year={2023},
  organization={PMLR}
}

@article{team2024octo,
  title={Octo: An open-source generalist robot policy},
  author={Team, Octo Model and Ghosh, Dibya and Walke, Homer and Pertsch, Karl and Black, Kevin and Mees, Oier and Dasari, Sudeep and Hejna, Joey and Kreiman, Tobias and Xu, Charles and others},
  journal={arXiv preprint arXiv:2405.12213},
  year={2024}
}

@article{hu2024data,
  title={Data scaling laws in imitation learning for robotic manipulation},
  author={Hu, Yingdong and Lin, Fanqi and Sheng, Pingyue and Wen, Chuan and You, Jiacheng and Gao, Yang},
  journal={arXiv preprint arXiv:2410.18647},
  year={2024}
}

@article{gao2024efficient,
  title={Efficient data collection for robotic manipulation via compositional generalization},
  author={Gao, Jensen and Xie, Annie and Xiao, Ted and Finn, Chelsea and Sadigh, Dorsa},
  journal={arXiv preprint arXiv:2403.05110},
  year={2024}
}

@inproceedings{xie2024decomposing,
  title={Decomposing the generalization gap in imitation learning for visual robotic manipulation},
  author={Xie, Annie and Lee, Lisa and Xiao, Ted and Finn, Chelsea},
  booktitle={2024 IEEE International Conference on Robotics and Automation (ICRA)},
  pages={3153--3160},
  year={2024},
  organization={IEEE}
}

@article{xu2024flow,
  title={Flow as the cross-domain manipulation interface},
  author={Xu, Mengda and Xu, Zhenjia and Xu, Yinghao and Chi, Cheng and Wetzstein, Gordon and Veloso, Manuela and Song, Shuran},
  journal={arXiv preprint arXiv:2407.15208},
  year={2024}
}

@misc{xu2025dexumiusinghumanhand,
      title={DexUMI: Using Human Hand as the Universal Manipulation Interface for Dexterous Manipulation}, 
      author={Mengda Xu and Han Zhang and Yifan Hou and Zhenjia Xu and Linxi Fan and Manuela Veloso and Shuran Song},
      year={2025},
      eprint={2505.21864},
      archivePrefix={arXiv},
      primaryClass={cs.RO},
      url={https://arxiv.org/abs/2505.21864}, 
}

\end{document}